\documentclass[10pt]{article}

\usepackage{epsfig}
\usepackage{amsmath}
\usepackage{amssymb}
\usepackage{color}
\usepackage{url}

\usepackage{verbatim} 
\usepackage{epstopdf} 
\usepackage{graphicx} 
\usepackage[font=small]{caption} 

\usepackage{afterpage}

\usepackage{caption}
\usepackage{subcaption}
\usepackage{multicol}

\usepackage{multirow}

\usepackage{setspace}

\usepackage[numbers]{natbib}

\usepackage{algorithm}
\usepackage{algorithmic}
\usepackage{mathrsfs}

\usepackage{multirow}

\newcommand{\commentout}[1]{}

\newcommand{\R}{\mathbb{R}}

\renewcommand{\arraystretch}{1.2}

\begin{document}

\title{Spatio-Temporal Surrogates for Interaction of a Jet with High
  Explosives: Part II - Clustering Extremely High-Dimensional
  Grid-Based Data}

\author{%
Chandrika Kamath and Juliette S. Franzman\\[0.5em]
{\small\begin{minipage}{\linewidth}\begin{center}
\begin{tabular}{c}
Lawrence Livermore National Laboratory \\
7000 East Avenue, Livermore, CA 94551, USA\\
\url{kamath2, franzman1@llnl.gov}\\
\hspace*{0.8in}
\end{tabular}
\end{center}\end{minipage}}
}

\date{9 June 2023}
\maketitle

\begin{abstract}
  Building an accurate surrogate model for the spatio-temporal outputs
  of a computer simulation is a challenging task. A simple approach to
  improve the accuracy of the surrogate is to cluster the outputs
  based on similarity and build a separate surrogate model for each
  cluster. This clustering is relatively straightforward when the
  output at each time step is of moderate size. However, when the
  spatial domain is represented by a large number of grid points,
  numbering in the millions, the clustering of the data becomes more
  challenging. In this report, we consider output data from
  simulations of a jet interacting with high explosives. These data
  are available on spatial domains of different sizes, at grid points
  that vary in their spatial coordinates, and in a format that
  distributes the output across multiple files at each time step of
  the simulation. We first describe how we bring these data into a
  consistent format prior to clustering. Borrowing the idea of random
  projections from data mining, we reduce the dimension of our data by
  a factor of thousand, making it possible to use the iterative
  k-means method for clustering. We show how we can use the randomness
  of both the random projections, and the choice of initial centroids
  in k-means clustering, to determine the number of clusters in our
  data set. Our approach makes clustering of extremely high
  dimensional data tractable, generating meaningful cluster
  assignments for our problem, despite the approximation introduced in
  the random projections.

\end{abstract}

\pagebreak
\begin{spacing}{0.7}
\tableofcontents
\end{spacing}

\pagebreak

%
\section{Introduction}
%

A common task in building surrogates for the spatio-temporal outputs
from computer simulations is to transform these outputs into a
lower-dimensional space using a linear method such as the principal
component analysis (PCA)~\cite{jolliffe2016:pcareview}. However, when
the data do not lie on a linear manifold, the dimension of this
lower-dimensional space can be large, prompting the consideration of
non-linear dimension-reduction methods. A simple approach to
introducing non-linearity is to cluster the simulation outputs by
similarity and then use a linear transform on each cluster
separately~\cite{kambhatla1997:localpca}, creating a locally-linear
surrogate. However, clustering the outputs becomes challenging when
the spatial domains vary across simulations or the spatial data
generated at each time step of the simulation is extremely large,
composed of values at over a million grid points in the spatial
domain. This report describes our experiences with addressing these
challenges.

We start this report by describing the problem for which we want to
build a spatio-temporal surrogate, namely, the interaction of a jet
with high explosives, and the two-dimensional outputs that are
generated during the simulations of this problem
(Section~\ref{sec:data_desc}). These outputs form the data set that we
want to cluster.  We then discuss the issues that make it challenging
to cluster these outputs (Section~\ref{sec:challenges}) and place our
contributions in the context of related work
(Section~\ref{sec:related}). Our solution approach
(Section~\ref{sec:approach}) describes how ideas from other domains
make it possible to cluster the high-dimensional data.  We show how we
can select the parameters used in our algorithms to generate
meaningful clustering results for our data set
(Section~\ref{sec:results}). We conclude this report with a summary of
our experiences and the lessons learned (Section~\ref{sec:conc}).

This report is the second of two reports summarizing our work on
building spatio-temporal surrogates for the problem of a jet
interacting with high explosives. In the first
report~\cite{kamath2023:stmapps}, we discuss the applications aspect
of our work and describe how we can build an accurate surrogate
despite the small number of simulations in our data set. One of the
approaches to improving the accuracy of the surrogate is by building
locally-linear surrogates. This requires clustering of the data, which
is the focus of this report.

%
\section{Description of the data}
\label{sec:data_desc}
%

\begin{figure}[!b]
\centering
\begin{tabular}{c}
\includegraphics[trim = 0.0cm 5.5cm 0.0cm 6.5cm, clip = true,width=0.65\textwidth]{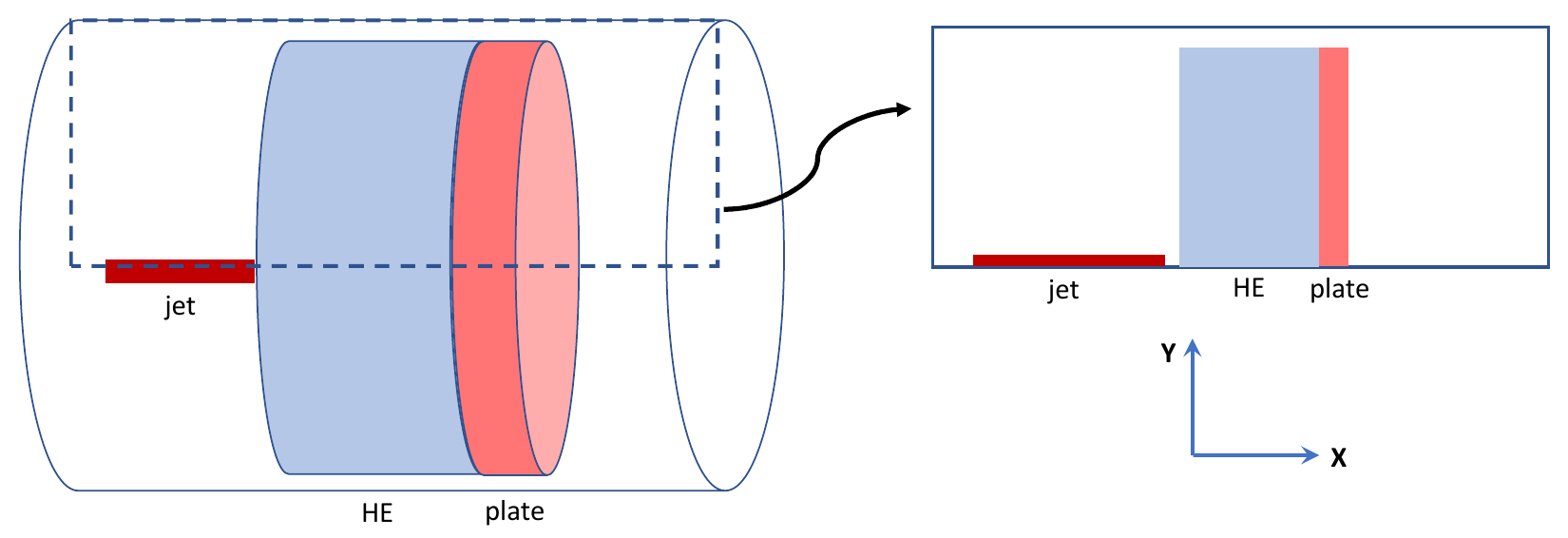}\\
\end{tabular}
\caption{A schematic of the problem being simulated. On the left is
  the horizontal cylinder in three-dimensions, showing the plate in
  pink, to the left of which is the HE in blue. The jet, in red,
  enters the HE from the left. As the problem is radially symmetric
  about the axis of the cylinder, we need to simulate only the
  two-dimension region shown by a dotted rectangle on the left and
  schematically on the right.}
\label{fig:schematic}
\end{figure}

We illustrate our ideas on clustering extremely high-dimensional data
using simulation output from a problem describing the interaction of a
jet with high explosives (HE). The domain of the problem is a right
cylinder with its axis oriented horizontally as shown on the left in
Figure~\ref{fig:schematic}. There is a steel plate, 1cm thick near the
right end of the cylinder, with the LX14 high explosives to the left
of the plate. Both the plate and the HE have a fixed radius of 10cm. A
copper jet, aligned along the center line of the cylinder, enters the
HE from the left. The simulation models what happens as the jet moves
through the HE and the plate. The jet is modeled initially as uniform
cylinder. It is 10 cm in length with a varying radius. The jet tip
velocity is specified as an input parameter;   a linearly varying
velocity profile is applied to the remainder of the cylinder that represents
the jet to approximate a stretching metal jet.  As the problem is
radially symmetric about the axis of the cylinder, only the
two-dimensional region shown by a dotted rectangle on the left, and
schematically on the right, is simulated. At each time step, the
simulation outputs variables of interest, such as mass and momentum,
at different points on a grid in the two-dimensional region.

There are three input parameters for the simulation: the {\it radius}
of the jet, the {\it length} of the high explosives to be traversed by
the jet, and the {\it tip velocity} of the jet in the positive $x$
direction.  By running the simulations at select values of these input
parameters, and collecting the output at different time steps, we can
create a data set that could be used to build a surrogate model for
predicting the output at a new set of input parameters and a given
time step. We are interested in determining, for example, whether the
plate breaks; what is the final position of the plate; and, if the
plate breaks, what is the velocity of the tip of the jet as it comes
out on the other side of plate. To answer these questions, we need to
build an accurate spatio-temporal surrogate, a topic we discuss in the
companion report~\cite{kamath2023:stmapps}, which focuses on the
accuracy of the surrogates created using only a small number of
simulations. This report focuses on the tasks of processing and
clustering of extremely high-dimensional data that are crucial to
building this accurate surrogate.

\begin{figure}[!htb]
\renewcommand{\arraystretch}{1.2}

\vspace{1cm}

\centering
\setlength\tabcolsep{1pt}
\begin{tabular}{cc}
\includegraphics[width=0.45\textwidth]{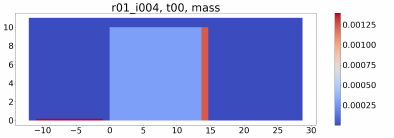} &
\includegraphics[width=0.45\textwidth]{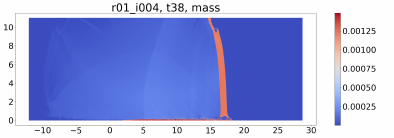} \\
\includegraphics[width=0.45\textwidth]{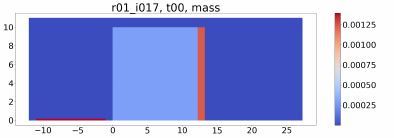} & 
\includegraphics[width=0.45\textwidth]{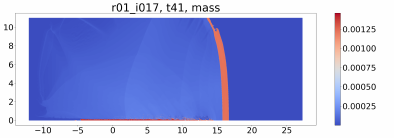} \\
\includegraphics[width=0.45\textwidth]{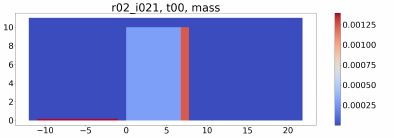} & 
\includegraphics[width=0.45\textwidth]{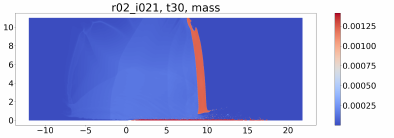} \\
\includegraphics[width=0.45\textwidth]{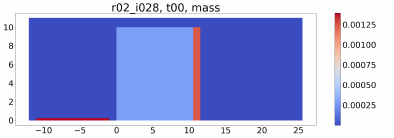} & 
\includegraphics[width=0.45\textwidth]{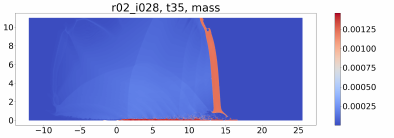} \\
\end{tabular}
\captionof{figure}{The variable {\it mass} at the first time step,
  (left column) and last time step (right column) for the four example
  simulations in Table~\ref{tab:sample_params}. From top to bottom,
  simulations with keys r01\_i004, r01\_i017, r02\_i021, r02\_i028,
  illustrate an {\it almost break}, {\it no break}, {\it break}, and
  {\it break} case, respectively.  The data shown are before the
  pre-processing steps described in Section~\ref{sec:preprocessing}.
  The vertical red region represents the plate, the light blue
  represents the high explosive (HE), the horizontal red region at the
  bottom to the left of the HE at the first time step is the jet, and
  the dark blue region is air. The right column shows the last time
  step of each simulation and the effect on the plate as the jet moves
  to the right, through the HE, and potentially through the plate. The
  range of x values is different for different simulations, while the
  range of y values is the same. The HE starts at $x=0$ in each plot.
  Note that even when the plate does not break (second row), the plate
  has moved to the right from its original position.}
\label{fig:sample_snaps_nmass}
\end{figure}

To illustrate the instances in our data set, we use four simulations
whose parameters are listed in Table~\ref{tab:sample_params}.
Figure~\ref{fig:sample_snaps_nmass} shows the output variable, {\it
  mass}, at the first and last time steps for these four example
simulations. As explained earlier, we have simplified the
three-dimensional problem by assuming radial symmetry around the axis
of the cylinder, so the output from the simulation is shown as two
dimensional images, with the axis of the cylinder shown at the bottom,
that is, at $y = 0$.  The domain extent in $x$ (along the length of
the cylinder) varies as the length of the HE varies across
simulations; however, the domain in $y$ ranges from 0 to 11cm for all
simulations.

\begin{table}[!htb]
  \begin{center}
    \begin{tabular}{|l|l|l|l|l|l|l|}
      \hline
      Simulation key & jet radius & HE-length & jet tip velocity & \texttt{\#} time & \texttt{\#}grid & outcome \\
                     & (cm)   & (cm)      & (cm/$\mu$sec) & steps & points & \\
      \hline
      r01\_i004 & 0.15 & 13.67 & 0.894 & 38 & 2,859,387 & almost break \\
      r01\_i017 & 0.17 & 12.24 & 0.648 & 41 & 2,759,181 & no break \\
      r02\_i021 & 0.14 & 6.77 & 0.914  & 30 & 2,374,179 & break \\
      r02\_i028 & 0.23 & 10.54 & 0.843 & 35 & 2,639,637 & break \\
      \hline
    \end{tabular}
  \end{center}
  \vspace{-0.2cm}
  \captionof{table}{Input parameters for the four example simulations shown in 
    Figures~\ref{fig:sample_snaps_nmass}. Note the very large number of 
    grid points (over two million) at which variables of interest are 
    output at each time step in a simulation.}
  \label{tab:sample_params}
\end{table}

In Figure~\ref{fig:sample_snaps_nmass} the vertical plate, shown in
red, is stationary at time $t = 0$. To the left of the plate is the HE
shown in light blue.  The jet is shown in red at the bottom of the
domain to the left of the HE; it is quite thin relative to the radius
of the cylinder, and is barely visible in the images.  As the
simulation evolves, the jet moves to the right, through the HE, which
expands, pushing the plate to the right. At late time, depending on
the simulation input parameters, the plate could 
\begin{itemize}

\item{\bf break}, with the jet going through the plate and 
  coming out clearly on the other side;

\item{\bf almost break}, with the jet either going completely through
  the plate but barely coming out the other side or the jet going
  almost all the way through the plate, leaving it barely connected at
  the bottom;

\item{\bf not break}, with the plate remaining attached, either
  partially or completely, at the bottom. The plate could have moved
  from its original position at time $t =0$.

\end{itemize}  
We used the last two time steps in each simulation to assign one of
these three class labels to the simulation. This label was
not used in building the surrogate; it was used only to ensure we had
a good coverage of the design space. We selected the four example
simulations in Figure~\ref{fig:sample_snaps_nmass} to illustrate these
three cases.

The output at each time step of a simulation consists of the values of
variables of interest that are generated at grid points in the
two-dimensional rectangular domain. These grid points are on a regular
grid, with $\Delta x = \Delta y = 0.0125$cm.  There are three output
variables: {\it mass, x-momentum, and y-momentum}; the latter two are
shown in later in Appendix~\ref{sec:appendix1}
and~\ref{sec:appendix2}, respectively.  The values of these variables
are defined at the center of the square cell formed by four nearby
grid points.  Thus the data appear as an image, with regularly spaced
pixels.  However, in general, in a simulation, the grid points need
not be on a regular grid; they could form an unstructured grid, as in
a finite element mesh, or a locally structured grid, as in an Adaptive
Mesh Refinement (AMR) mesh, where the mesh evolves with the
simulation.  As a result, unlike an image, most output from
simulations also include the $(x,y)$ coordinates of the grid points.
In our work, we retain this association of the coordinates with the
grid points as they enable us to extract sub-domains of the larger
domain for processing.

Each simulation is run for a fixed number of time steps which is
determined as $( \lfloor (\text{HE-length}/\text{jet-tip-velocity} \rfloor ) + 23
)$, with the output generated at each time step. As both HE-length and
jet tip velocity vary with the simulation, the number of time steps also varies
across simulations. At early time, as the jet starts to move through
the HE, there is little of interest in the simulation output. Once the
jet is partway through the HE, as indicated by the first term in the
equation above, it starts to influence the location of the plate,
until 23 $\mu$sec later, it is expected that we should know the final
status of the plate. In our work, we consider all the time steps in
the analysis; an alternative would be to consider only the later 23
time steps.

\begin{figure}[htb]
\centering
\begin{tabular}{cc}
\includegraphics[trim = 0cm 0cm 0cm 0.0cm, clip = true,width=0.5\textwidth]{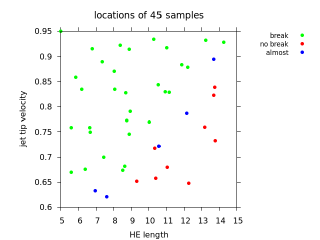} &
\includegraphics[trim = 0cm 0cm 0cm 0.0cm, clip = true,width=0.5\textwidth]{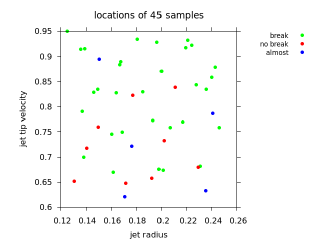} \\
\end{tabular}
\caption{The 45 samples in the space of the three input parameters for
  the simulation, labeled by the state of the plate at the last time
  step of the simulation. The simulation at the extreme corner of the
  input space, with length, jet tip velocity, and jet radius equal to
  5.0cm, 0.950cm/$\mu$s, and 0.125cm, respectively, is referred to as
  the {\it baseline} simulation. It has the smallest number of time
  steps. Our goal is to cluster all the time steps across these 45
  simulations.}
\label{fig:all_samples}
\end{figure}

The data processed in this report was obtained by running 45
simulations at select values of the input parameters.  These values,
or sample points in the three input dimensions, were generated
incrementally using a modified version of the best candidate
algorithm~\cite{mitchell91:sampling,kamath2022:sampling}, which selects
samples randomly, but far apart from each other. We started with a
small number of sample points and an initial guess at the range of
each of the three inputs. We then restricted the ranges to focus on
the break cases, and added new sample points until we had a total 45
sample points~\cite{kamath2023:stmapps}. Our data set, shown in
Figure~\ref{fig:all_samples}, indicates that at high jet tip velocity, but
low HE-length, the plate breaks, while at low jet tip velocity and high
HE-length, the jet is unable to penetrate through the plate.

It is clear that our data set is unbalanced as we have 9 samples where
the plate does not break, 31 where the plate breaks, and 5 that are
almost break.  Generating an appropriate data set for a problem like
ours is challenging as time constraints limit the number of
simulations we can run. In addition, we do not know {\it a priori} the
range of inputs that will give us sample points with the desired
outcome, and the boundary between the classes is poorly defined,
making it difficult to generate a balanced data set.  We erred on the
side of having more break cases as these were of greater interest; the
no-break cases also tended to have output that appeared very similar,
and we expected that a small number of such cases would suffice.
Admittedly, our choice of sample points will affect the clustering
results.

The output data for a variable at a time step in a simulation is
referred to as a {\it snapshot}, so named as it is a snapshot of the
evolution of the simulation at a particular point in time.  These 45
simulations generate a total of 1604 snapshots.  The simulation at the
extreme corner of the input space, with length, jet tip velocity, and jet
radius equal to 5.0cm, 0.950cm/$\mu$s, and 0.125cm, respectively, is
referred to as the {\it baseline} simulation. It has the smallest
number of time steps, with 29 snapshots.

%
\section{Challenges to the analysis}
\label{sec:challenges}
%

There are two main challenges to building a high-quality spatio-temporal
surrogate for our problem:

\begin{itemize}

\item The first is how do we build a surrogate that is {\it accurate}
  when we can only run a small number of simulations? We discuss
  several options in our companion report~\cite{kamath2023:stmapps},
  one of which is to cluster the snapshots and build a separate
  surrogate for each cluster.

\item This leads to our second challenge - how do we cluster the
  snapshots in our data set? Table~\ref{tab:sample_params} and
  Figure~\ref{fig:sample_snaps_nmass} indicate that the snapshots are
  high-dimensional, each with more than two million grid points; the
  sizes of the snapshots vary across simulations; and the snapshots
  are not aligned in any way; and the data at each time step are
  available in multiple files. All these factors would make clustering
  the snapshots challenging. In addition, popular clustering
  algorithms, such as k-means, are iterative, which can lead to
  computational inefficiencies in processing extremely high-dimensional
  data.

\end{itemize}

The first of these two challenges is the focus of the companion
report~\cite{kamath2023:stmapps}; this report focuses on the second
challenge, namely, generating a meaningful clustering of the
high-dimensional snapshots from our problem of jet-HE interaction.  To
accomplish this goal, we need to address the two issues discussed
next.

%
\subsection{The unsuitability of the raw data for clustering}
\label{sec:challenge1}
%

The output data generated for each of the 45 simulations, regardless
of the size of the domain, are available in 360 files in HDF5
format~\cite{hdf5:online} for each time step.  Each HDF5 file includes
five variables --- $x$ and $y$ coordinates, mass, x-momentum, and
y-momentum. Within each file, the variables are in natural ordering,
that is, ordered by increasing values of the $y$-coordinate, and for a
fixed $y$-coordinate, ordered by increasing values of the
$x$-coordinate.  All simulations are on a regular grid with $\Delta x
= \Delta y = 0.0125$cm.

\begin{figure}[!htb]
\centering
\begin{tabular}{c}
\includegraphics[trim = 0.0cm 0cm 0cm 0.0cm, clip = true,width=0.75\textwidth]{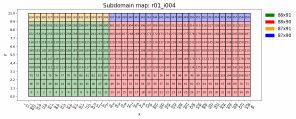} \\
\includegraphics[trim = 0.0cm 0cm 0cm 0.0cm, clip = true,width=0.75\textwidth]{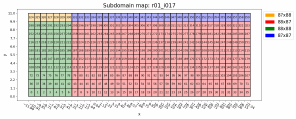} \\
\includegraphics[trim = 0.0cm 0cm 0cm 0.0cm, clip = true,width=0.75\textwidth]{r01_i004_subdomain_plot.pdf} \\
\includegraphics[trim = 0.0cm 0cm 0cm 0.0cm, clip = true,width=0.75\textwidth]{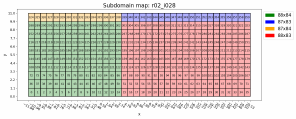} \\
\end{tabular}
\caption{The distribution of data in 360 sub-domains for the four
  example simulations in Table~\ref{tab:sample_params}. At each time
  step, the data in each sub-domain is available in a separate HDF5
  file. From top to bottom, simulations with keys r01\_i004,
  r01\_i017, r02\_i021, r02\_i028, respectively.  Note that the sizes
  of the sub-domains within a simulation vary. All time steps in a
  simulation have the same distribution of the data.}
\label{fig:sample_subdomains}
\end{figure}

To apply our analysis algorithms to these data, we first need to
re-arrange the data so that we have three snapshot matrices, one for
each of the three variables of interest. Each snapshot matrix should
have the grid points as the rows, listed in natural order of the
$(x,y)$ coordinates, and, as columns, the variable values at each time
step of each simulation. A snapshot matrix, ${\bf X} \in {\R}^{D
  \times N}$, for any one of the three variables, can then be written as
\begin{equation}
{\bf X} = \left[ {\bf x}_1, {\bf x}_2, \ldots, {\bf x}_N \right] ,
\label{eqn:smatrix}
\end{equation}
where ${\bf x}_i \in \R^D$, $N$ is the number of snapshots, and $D$ is
the number of grid points in a snapshot.  Since we have a total of
1604 time steps in the 45 simulations, there are 1604 columns in the
snapshot matrix. However, identifying the rows of the snapshot
matrix is more challenging. To cluster the snapshots, we
first need to bring the data in each simulation to a common grid,
which is not straightforward for several reasons:
\begin{itemize}

\item The domain over which the data are generated is different for
  each simulation. The $y$ values are in the range [0:11.0] cm for all
  simulations, but the range of $x$ values varies as the HE-length
  varies, resulting in snapshots that vary in length across
  simulations. 

\item Figure~\ref{fig:sample_snaps_nmass} shows that the origin of the
  domain in all simulations is at the left edge of the HE, with 12 cm
  of air to the left of the origin. 
  So, if we align the snapshots at the left edge at $x = -12.0$, the
  plate locations will not be aligned, even at time step 0, as the HE
  length varies across simulations. Since the plate forms an important
  structure in the output data, the lack of alignment of the plate
  across simulations would not give us the clustering results we
  expect.

\item For a simulation, the distribution of the data in the 360 files
  is the same across all time steps. However, the files within a
  simulation have four different sizes, as shown in
  Figure~\ref{fig:sample_subdomains}. But, as the domain size
  varies across simulations, the sizes of these HDF5 files also vary
  across simulations. As each row in the snapshot matrix corresponds
  to the same $(x,y)$ coordinates, it becomes non-trivial to map a
  point in one of the 360 files at a time step to its location in
  the snapshot matrix efficiently.

\item All the simulations have the coordinate $(-12,0)$ as the lower
  left corner. But, due to the way in which the $(x, y)$ coordinates
  are generated for domains with different ranges in $x$, the values
  of $\Delta x$ and $\Delta y$ are not exactly 0.0125cm across
  simulations.  While the coordinates in $y$, which has a fixed
  range of values, are identical across simulations, this is not the
  case for the $x$ coordinates.

\item Finally, the total number of grid points across the 360 HDF5 files for
  the data at any time step is nearly three million (see
  Table~\ref{tab:sample_params}).  A snapshot matrix with such a large
  number of rows and 1604 columns is too large to be read into memory
  for processing and we need to consider alternative ways to store the
  snapshot matrix; this would also affect the analysis algorithms that
  read in the data.

\end{itemize}

\afterpage{\clearpage}

%
\subsection{The iterative nature of k-means clustering algorithm}
\label{sec:challenge2}
%

In addition to the challenges posed by the size of the raw data and
the storage format, we also need to consider the clustering algorithm
used to group the snapshots. We intuitively expect that the snapshots
can be clustered because the early time steps, when the plate has
barely moved, are distinctly different from the later time steps,
where, depending on the input parameters of the simulation, the plate
moves but does not break, or the plate moves and breaks. However, as
neighboring time steps are often quite similar, any clustering will
place some consecutive time steps from a simulation into different
clusters, suggesting that any clusters identified are not very well
separated.

In the absence of any information on the shape of possible clusters or
the density of points in the very high-dimensional space, it is not
obvious which  clustering algorithm is the most suitable one for our data.
We chose to start with the simplest clustering method, k-means
clustering~\cite{jain88:clusbook,gan2007:clusbook},
described in Algorithm~\ref{algo:kmeans}, with the expectation that it
might provide insight to guide the choice of a more appropriate
algorithm.

\begin{algorithm*}[htb]
  \caption{$k$-means clustering }
\label{algo:kmeans}
\begin{algorithmic}[1]

  \STATE Goal: given the $N$ snapshots in a $D$-dimensional space in
  the form of the matrix $X$ in Equation~\ref{eqn:smatrix}, identify a
  clustering with $nc$ clusters, such that each snapshot is assigned to
  a cluster, and the snapshot is closer to the centroid of its cluster
  than the centroid of any other cluster.
  
  \STATE Set $ niter$, the maximum number of iterations and $thresh$,
  the threshold that determines when a cluster center has moved too
  small an amount between iterations.

  \STATE Choose an initial set of $nc$ snapshots randomly as the
  initial cluster centroids.  

  \FOR{$iter = 1$ to $niter$}

  \STATE Assign each snapshot to its nearest cluster centroid using
  the Euclidean distance. 

  \STATE Update each cluster centroid to be the mean of the snapshots assigned to it. 

  \STATE If the maximum Euclidean distance moved by any centroid is
  less than $thresh$, terminate.

  \ENDFOR

\end{algorithmic}
\end{algorithm*}

We observe that at each iteration, this algorithm requires the
calculation of the distance of each snapshot to the centroid of each
cluster. As the snapshot matrix is too large to fit into memory,
reading in the matrix, piece-meal, on each iteration, will be time
consuming.  We therefore need alternate ways to make the clustering
of the snapshots computationally tractable.

%
\section{Related work}
\label{sec:related}
%

We address two tasks in this report - converting the data from the HDF5
files into a snapshot matrix and clustering the columns of the
high-dimensional snapshot matrix. The first task is specific to each
data set and is based on the characteristics of the data.  This
section therefore focuses on the second task.

In the field of spatio-temporal modeling, the problems solved 
typically have snapshots with grid points numbering in the thousands
or tens of thousands, and occasionally hundreds of
thousands~\cite{kamath2023:stmapps}.  Clustering these relatively
low-dimensional snapshots then becomes a straightforward application
of a clustering algorithm, such as k-means or k-mediods, with a
suitably defined distance metric, such as Euclidean distance,
reconstruction distance, or Grassmann
distance~\cite{kambhatla1997:localpca,amsallem2012:nonlocal,du2002:podcvt,daniel2020:romnet}.

To cluster extremely high-dimensional data, where each snapshot has
over two million grid points, we borrowed ideas from the field of data
mining, where such data sets are common place. A typical solution is
to first reduce the dimension and then cluster the data.  Ritter and
Kohonen in 1989 suggested using random
projections~\cite{achlioptas2003:randproj} for dimension reduction,
followed by clustering using self-organizing maps, which is another
iterative technique like k-means. Using the same combination of
methods, Kaski in 1998~\cite{kaski1998:randproj} showed that random
projections was faster than PCA for dimension reduction; it required a
slightly larger number of reduced dimensions, but the results were
equally good, and similar to the results with the original data.
In later work, Fern and Brodley~\cite{fern2003:randproj} found that
the randomness of random projections could result in different cluster
assignments and proposed using ensemble
clustering~\cite{strehl2002:ensemble} to generate a stable cluster
assignment. They also found that random projections performed better
than PCA, and that there was no universal single best way to combine
the results of the ensemble~\cite{fern2004:cluster}. More recently,
Anderlucci et al.~\cite{anderlucci2022:rpclus} have further
investigated different ways to combine the different cluster
assignments resulting from random projections.

Based on this prior work, we selected random projections as the
approach for reducing the dimension of the three snapshot matrices in
our data set. Our contributions in this work are as follows:

\begin{itemize}

\item We show how we can generate the snapshots matrices when the data 
  available are on spatial domains with different sizes, at grid points
  that vary in their $(x,y)$ locations across simulations, and in a format that
  distributes the output across multiple files at each time step of a simulation.

\item We discuss how we can apply random projections to extremely
  high-dimensional data, determine the reduced dimension, and evaluate
  the results.

\item We indicate how we can exploit the randomness resulting both from the random
  projections, and the initial choice of centroids in k-means
  clustering, to obtain the number of clusters in the data.

\end{itemize}

%
\section{Solution approach}
\label{sec:approach}
%

Our approach to clustering the very-high dimensional data set
resulting from the simulations of a jet interacting with high
explosives (HE) requires us first to convert the raw data into a
format that we can cluster and then to identify a computationally
feasible approach to clustering these high-dimensional snapshots. We
next describe how we address these tasks.

%
\subsection{Pre-processing the data}
\label{sec:preprocessing}
%

As described in Section~\ref{sec:challenge1}, we have data for 1604
time steps across the 45 simulations. For each simulation, at each
time step, the three variables, along with the corresponding $(x,y)$
coordinates of the grid points, are spread across 360 files, each
covering a small part of the domain.  We need to convert this to a
snapshot matrix for each variable, where the matrix has all the grid
points, in natural ordering, along the rows, and each column
contains the values of the variable at a different snapshot.

At first glance, this appears to be just a re-arrangement of data from
one set of files to another. But as we described in
Section~\ref{sec:challenge1}, we also need to account for the
different ranges of $x$ values, the lack of alignment of the plate
even at early time, and the inconsistent values for the $(x,y)$
coordinates across the simulations. This will require computation on
the data values, in addition to the re-arrangement of values in files.

We first decided that instead of storing the final data for each
variable in a single snapshot matrix file as in
Equation~\ref{eqn:smatrix}, we would split the matrix into blocks,
with each block stored in a separate file, as follows:
\begin{equation}
{\bf X} =
\begin{bmatrix}
{\bf X}_{b1} \\
{\bf X}_{b2} \\
\ldots \\
{\bf X}_{bk} \\
\end{bmatrix}
\label{eqn:blockx}
\end{equation}
This would make the processing of the very large number of grid points
tractable as we could read the matrix a block at a time. However, all
processing would have to be modified to work with the matrix spread
across several files.  We chose each block to contain all grid points
in a specific non-overlapping range of $y$ values. As the grid points
are stored in natural order within a block, concatenating the blocks
in the order of their $y$ values creates a single snapshot-matrix
file with all the grid points stored in natural order.

There are many ways in which we can transform the HDF5 data from 360
files per time step, per simulation, into the three snapshot matrices,
one for each variable that represents all the data for the variable
across the 45 simulations. The solution we describe next is one we
found expedient to implement. We used computationally efficient
implementations, exploiting parallelism where possible. As we wanted
our codes to remain flexible for processing other data formats, we
kept the steps in this multi-step process distinct, though for higher
computational efficiency, we could have merged some steps, or
executed them in a different order.

%
\subsubsection{Creating a data file for each time step in a simulation}
\label{sec:filepersim}
%

We started by first processing the raw data for each simulation. For
each time step in the simulation, the data are available in 360 files,
one for each sub-domain.  Each file contains the $(x,y)$ coordinates
and all variable values for that sub-domain.  In this first step, the
360 files were combined into a single file for each time step in a
simulation.  This essentially involved reading each file in turn,
extracting the coordinates and variable values, and writing out this
information, a row at a time, to the single data file.  The rows in
this file are the grid points in the same order as they appear in the
HDF5 file, starting at sub-domain 0 and ending at sub-domain 359. The
columns are the $(x,y)$ coordinates and the three variable values. For
each time step, we also generated a summary file with statistics on
each sub-domain, including the range of $x$ and $y$ values, and the
starting location of each sub-domain in the single data file. This
file is used to improve the computational efficiency of subsequent
steps in the processing of the data.

This step is just a re-arrangement of the data and involves reading a
set of files, extracting the relevant information, and writing it out.
The time steps across the simulations can all be processed in
parallel.

%
\subsubsection{Aligning and cropping coordinates to a common domain}
\label{sec:aligning}
%
 
After the first step, we have a large file for all variables at each
time step in a simulation, along with an associated summary file. When
we consider these files across the 45 simulations, the number of grid points
is different as the range of $x$ values is different for each
simulation.  In addition, as the origin in each simulation is at the
start of the HE, the plate is not aligned across simulations at time
t=0.

To fix this, we first changed the origin of the coordinates in each
file to be at the lower right corner of each domain.  This shift in
$x$ values is the same for all time steps in a simulation, but differs
across simulations. This step automatically aligns the plate at time
$t = 0$ across all simulations because the left edge of the 1cm wide
plate is 15 cm from the right edge of the domain. Such alignment of
the data is common in tasks such as face recognition using the
eigenface approach~\cite{turk1991:eigenfaces}, where each image is
pre-processed so the face covers the full image, is upright, and
centered in the image.  In our problem, unlike the face, which is
stationary, the plate and the HE move as the simulation progresses. By
aligning the plate at initial time we reduce the amount of
misalignment of the plate across the snapshots.

Next, we cropped the data in each file so that grid points with the
new shifted $x$ coordinate outside the range [-32, 0.0] are excluded.
This range was selected as it corresponds to the smallest HE length of
5.0cm. After this step, the data files for all simulations have the
same range of $x$ values and the same plate location at initial time.
However, the values of the variables are at different $(x,y)$
coordinate locations and the grid points are still in the same order
as they were within each sub-domain, with the sub-domains stored in
order.

These changes make it possible to generate a meaningful clustering of
the snapshots; they can be applied in parallel for the time steps
across the simulations.

%
\subsubsection{Remapping simulations to a common grid}
\label{sec:remapping}
%

Next, we used a simple interpolation scheme to remap the data values
for each simulation to a common grid. This remapping consists of two
steps. We first defined the common grid by choosing an $x$-range of
[-31.5, -0.5], which is slightly smaller than the range for each
simulation to ensure that all remapped values were being interpolated,
not extrapolated. The $y$-range was selected as [0.0063, 10.99]. We
kept the grid resolution the same at $\Delta x = \Delta y = 0.0125$,
resulting in a total of $D = 2,180,799$ grid points in the common
grid. Then, given the fine resolution of the common grid, we used a
simple 1-nearest-neighbor algorithm for interpolation, though more
complex algorithms are also an option.

We created the common grid in a block form, as in
Equation~\ref{eqn:blockx}, with each block written to a separate file.
A block spans the full range of $x$ values, but a smaller range of $y$
values. We chose $bk = 22$ blocks, with each block representing
approximately 0.5cm of the $y$-range. This gave approximately 100K
grid points in the first 21 blocks, and a smaller number in the last
block. Within a block, the grid points were listed in natural ordering
so that any data remapped to this common grid would have the rows in
the form required by the final snapshot matrix.

We generated the remapped data a block at a time.  For each time step
in a simulation, we extracted a block of data that had a range of
$(x,y)$ coordinates slightly larger than the block of the common grid
to which we were remapping this data. This ensured that we were
interpolating, not extrapolating, to the common grid. For
computational efficiency, we used the summary file associated with the
time step to identify and process only those sub-domains that had
coordinates in the required range.

We also observed that for our data, all time steps of a simulation
could be remapped together because the data extracted for a block for
the different time steps had values at the same $(x,y)$ coordinates in
the same order.  So, for each block, we first concatenated the columns at
different time steps to create a file for each simulation with
$(2+3*\#time steps)$ columns to store the $(x,y)$ coordinates and the
three outputs across all the time steps.  Merging these columns is
possible as we had maintained the ordering of the data in the
sub-domains and a fixed grid was used for all time steps. 

This re-mapping for the 45 simulations can be performed in parallel.
Once we have remapped the data from each simulation to each block of
the common grid, we have all data for the 45 simulations in the same
order as in the common grid, which simplifies the next steps
discussed in Section~\ref{sec:snapshot}

This step of remapping has several benefits. It introduces some
flexibility in the analysis as we can reduce the data size by mapping
to a coarser grid or remapping only the data in a smaller sub-region
of the full domain when the rest of the domain is of lesser interest.
By remapping to the common grid that is in the block form of the final
snapshot matrix, we can generate the remapped data for each block in
parallel, and with the smaller file sizes, process the data in
memory as well.  However, we observe that the computational
efficiencies in the remapping are the result of the same grid being
used across time steps in a simulation.  Generating the common grid
will be more challenging for AMR grids; the remapping will also be more
time consuming as it has to be done separately for each time step.

%
\subsubsection{Converting remapped data to snapshot matrices}
\label{sec:snapshot}
%

After the remapping step, we have 22 files for each simulation. Each
file corresponds to a block, with columns containing all time steps
for the three variables, along with the $(x,y)$ coordinates. The
remapping ensures that these coordinates are the same across the
simulations.

To create the final snapshot matrices, we split each block for a
simulation into three files, one for each variable, and then merge the
columns of these files across simulations. For each variable, this
gives the values at 2,180,799 grid points, split across 22 files, with
each file having 1604 columns, which is the total number of time steps
across the simulations. These snapshot matrices do not include the
$(x,y)$ coordinates.

\begin{figure}[htb]
\centering
\setlength\tabcolsep{1pt}
\begin{tabular}{cc}
\includegraphics[width=0.48\textwidth]{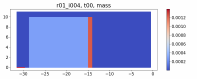} &
\includegraphics[width=0.48\textwidth]{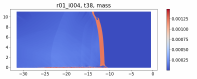} \\
\includegraphics[width=0.48\textwidth]{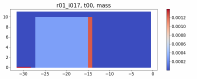} &
\includegraphics[width=0.48\textwidth]{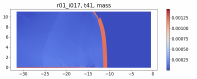} \\
\includegraphics[width=0.48\textwidth]{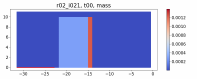} &
\includegraphics[width=0.48\textwidth]{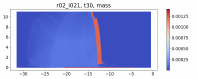} \\
\includegraphics[width=0.48\textwidth]{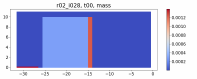} &
\includegraphics[width=0.48\textwidth]{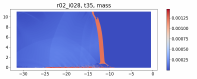} \\
\end{tabular}
\caption{The variable {\it mass} for the four example simulations in
  Table~\ref{tab:sample_params}, after the snapshots have been
  aligned, cropped, and remapped. Left column shows the first time
  step and right column shows the last time step. From top to bottom,
  simulations with keys r01\_i004, r01\_i017, r02\_i021, r02\_i028,
  respectively. The color bars are different between simulations and
  across time steps.}
\label{fig:sample_aligned_nmass}
\end{figure}

Figure~\ref{fig:sample_aligned_nmass} shows the mass variable for the
first and last snapshot, for each of our four example simulations,
after the raw output data have been aligned, cropped, and remapped.
The corresponding images for x-momentum and y-momentum are shown in
Figures~\ref{fig:sample_aligned_nxmom}
and~\ref{fig:sample_aligned_nymom} in Appendix~\ref{sec:appendix1}
and~\ref{sec:appendix2}, respectively.

%
\subsection{Clustering the snapshots}
\label{sec:clustering}
%

Once the three snapshot matrices, corresponding to the three output
variables, are created in block form, we can cluster the columns in
the matrices. In Section~\ref{sec:challenge2}, we described how the
high dimensionality of our data can be an issue for iterative
clustering algorithms as the full snapshot matrix is too large to fit
into memory. There are two obvious solutions to this problem of high
dimensionality that we discuss next.

%
\subsubsection{Using an iterative method with dimension reduction}
\label{sec:rpclustering}
%

Our first solution is to use the iterative k-means clustering
algorithm, but with modifications to account for the high
dimensionality of the data. One way to achieve this is by using a
distributed-memory version of the algorithm, which operates on data
divided into blocks (as in Equation~\ref{eqn:blockx}).  We can assign
each data block, and the corresponding block of the k-means centroids,
to a processor and calculate, in parallel, the partial
distance-squared of each snapshot to each centroid. Then a
communication step would add the partial distances and identify the
closest centroid to each snapshot, followed by an update of each block
of the centroids. This approach has limited parallelism equal to the
number of blocks and would require implementing a distributed-memory
version of the k-means algorithm.

Alternately, we could reduce the dimension of the data so that the
transformed matrix fits into the memory of a single processor. To
accomplish this, we used random
projections~\cite{achlioptas2001:dbfriendly,achlioptas2003:randproj}
which projects the $D$-dimensional matrix $X$ onto a $d$-dimensional
subspace, with $d \ll D$ using a random matrix $R \in {\R}^{d \times
  D}$

\begin{equation}
X^{RP}_{d \times N} = R_{d \times D} X_{D \times N}
\label{eqn:randproj}
\end{equation}

We can justify the use of random projections as the
Johnson-Lindenstrauss lemma~\cite{dasgupta2002:jlproof} states that
for any distortion $\epsilon$, with $0 < \epsilon < 1$, and any
integer $N$, if $d$ is a positive integer such that
\begin{equation}
d \ge 4 \left( \frac{\epsilon^2}{2} - \frac{\epsilon^3}{3} \right)^{-1} \ln N ,
\label{eqn:reddim}
\end{equation}
then, for any set $X$ of $N$ points in  $\R^{D}$, there is a map, $f : \R^{D} \rightarrow \R^{d}$ such that for all $u, v \in X$,
\begin{equation}
(1-\epsilon) \| u -v \|^2 \ \le \ \| f(u) - f(v) \|^2 \ \le \ (1+\epsilon) \| u -v \|^2 .
\label{eqn:jllemma}
\end{equation}
In other words, points in a sufficiently high-dimensional space can be
projected onto a suitable lower dimensional space, while approximately
preserving the distances between the points.  As the k-means algorithm
is based on Euclidean distances, this means we can cluster the
randomly-projected snapshot matrix $X^{RP}$ instead of the original
snapshot matrix $X$ and expect to get approximately similar clustering
results.

To apply the map $f$ in Equation~\ref{eqn:jllemma}, in the form of a
random projection of our snapshot matrix, $X$, we need to define the
random matrix, $R$, and determine the new reduced dimension, $d$.  As
our snapshot matrix is extremely high dimensional, we prefer a sparse
matrix, such as the one proposed by
Achlioptas~\cite{achlioptas2001:dbfriendly,achlioptas2003:randproj}
with i.i.d. entries as:
\begin{equation}
r_{i,j} = \sqrt{s}
\begin{cases}
1 & \text{with probability $\frac{1}{2s}$} \\
0 & \text{with probability $1-\frac{1}{s}$} \\
-1 & \text{with probability $\frac{1}{2s}$}
\end{cases}
\end{equation}
where $s = 1$ or $s = 3$, with the former resulting in a dense
projection matrix. Li, Hastie, and Church~\cite{li2006:randproj},
proposed using an even higher value of $s = {\sqrt{D}}$. For our
problem, with $D = 2,180,799$, this choice results in a very sparse
matrix, whose sparsity can be exploited for computational efficiency.
As the random projection changes the distances between two snapshots
by $(1 \pm \epsilon)$, we can obtain the new dimension $d$ by first
choosing a value of $\epsilon$, that is, the distortion we can
tolerate, and use $N = 1604$ in Equation~\ref{eqn:reddim}. For
$\epsilon = 0.1$, we get $d > 6325$, while $\epsilon = 0.05$ gives $d
\ge 24431$ and $\epsilon = 0.01$ gives $d \ge 594383$.

It has been observed that values of $d$ less than the ones suggested
by Equation~\ref{eqn:reddim} work well in
practice~\cite{bingham2001:randproj}. We chose $d$ based on both the
distortion in distances we can tolerate and the space available to
store the randomly-projected snapshot matrix $X^{RP}$ in memory. We
generated $X^{RP}$ incrementally by reading in each block of the
snapshot matrix $X$ a row at a time, generating a column of the sparse
random matrix $R$, and calculating the outer product of the column and
row and adding it to $X^{RP}$.
By not storing either $R$ or a block of $X$, we can choose a larger
value for $d$ as we require storage only for $X^{RP}$.

Once we have obtained the randomly projected snapshot matrix $X^{RP}$
of size $d \times 1604$, we can apply the k-means algorithm to cluster
the randomly-projected snapshots.  The clustering results so obtained
could vary based on the randomness of the projection and the random
initial choice of the cluster centroids.  We discuss these issues, and
our choice of parameters for random projections and k-means, further in
Section~\ref{sec:results}.

%
\subsubsection{Using an iterative method with a reduced representation}
\label{sec:wtclustering}
%

An alternative that is similar in spirit to the use of random
projections, but specific to our approach to building spatio-temporal
surrogates, is to exploit an intermediate step in the building of
these surrogates. As explained in the companion
report~\cite{kamath2023:stmapps}, we perform a singular value
decomposition (SVD) on the snapshot matrix ${\bf X}$ from
Equation~\ref{eqn:smatrix} to obtain
\begin{equation}
{\bf X} = {\bf U}  {\bf \Sigma}  {\bf V}^T 
\label{eqn:svd}
\end{equation}
where 
\begin{equation} \notag
{\bf X} \in \R^{D \times N}
\quad , \quad
{\bf U} \in \R^{D \times D}
\quad , \quad
{\bf \Sigma} \in \R^{D \times N}
\quad , \quad \text{and} \quad
{\bf V} \in \R^{N \times N} .
\end{equation}
Here, ${\bf \Sigma}$ is a diagonal matrix whose non-zero diagonal
elements, $\sigma_{ii}$, are the singular values of the matrix $\bf X$
and the columns of the orthonormal matrices ${\bf U}$ and ${\bf V}$
are the left and right singular vectors of $\bf X$.  The rows and the
columns of the ${\bf U}$ and ${\bf V}$ matrices are ordered such that
the singular values are in descending order in ${\bf \Sigma}$.  Since
$D >> N$, only the top $N$ rows of ${\bf \Sigma}$ will have non-zero
diagonal elements, assuming rank($\bf X$) = $N$. If the rank, $r$, is
less than $N$, then only the first $r$ diagonal elements are non-zero.
The columns of the matrix
\begin{equation}
{\bf U} = \left[ {\bf u}_1, {\bf u}_2, \ldots, {\bf u}_N \right] \quad \text{where} \quad {\bf u}_i \in \R^D
\end{equation}
form a basis in $\R^D$ for the data, so each snapshot
can be written as a linear combination of the ${\bf u}_i$:
\begin{equation}
{\bf x}_i = \sum_{k=1}^N w_{ki} {\bf u}_k
\label{eqn:reconstruct_n}
\end{equation}
where  the coefficient $w_{ki}$ of the $k$-th basis for the $i$-th snapshot is
\begin{equation}
w_{ki} = {\bf u}_k^T {\bf x}_i \quad \text{for} \quad k = 1, \ldots, N .
\label{eqn:coeffs}
\end{equation}
In building the spatio-temporal surrogate, we consider only the more
important ${\bf u}_i$ corresponding to the larger singular values,
and, by truncating the summation in Equation~\ref{eqn:reconstruct_n},
we create a reduced representation.
However, for the purpose of clustering the snapshots,
we can view the reduced representation of the $i$-th snapshot as the
vector of the $N$ weights
\begin{equation}
{\bf w}_i = \left[ w_{1i},  w_{2i}, \ldots,  w_{Ni} \right] ,
\end{equation}
and then obtain a clustering of the snapshots by clustering the
vectors of weights that define each snapshot. If only the few initial
weights are calculated, they can be used as approximate
representations of a snapshot.

In efforts where we want to compare the quality of a global
spatio-temporal surrogate, built using all snapshots, with a set of
local surrogates created after clustering the snapshots, we can
generate the cluster assignment directly using the weights obtained
from the SVD of the full data set (Equations~\ref{eqn:svd}
and~\ref{eqn:coeffs}), avoiding any additional computation.  However,
if we are creating only the local surrogates, this approach would be
computationally more expensive than using random projections with
sparse matrices (Section~\ref{sec:rpclustering}).

In this report, we considered the option of clustering the snapshot
using the weights obtained from the SVD as it allows us to compare the
cluster assignment obtained from an exact representation of the data
with that obtained from an approximate representation generated by
random projections.

%
\subsubsection{Using a non-iterative method}
\label{sec:hclustering}
%

An alternative to using an iterative clustering algorithm, which is
made tractable using distributed memory processing or a dimension
reduction technique, is a clustering algorithm that does not require
iterating over the snapshot matrix. One such method is hierarchical
clustering~\cite{jain88:clusbook,gan2007:clusbook}, a greedy algorithm
which, in its agglomerative form, begins with every data point in its
own cluster. At each step a linkage criterion is used to measure the
similarity between all the clusters, and the two most similar clusters
are merged together. This process of merging clusters continues until
the desired number of clusters are obtained or until the clusters no
longer satisfy a desired level of similarity
\cite{ward1963:clustering}. Hierarchical clustering is usually
accompanied by a dendrogram, so this stopping condition can be
visualized as making a cut across the dendrogram at the specified
similarity threshold to create a set of clusters.  Hierarchical
clustering requires the following parameters: a distance metric, often
Euclidean; a linkage criterion; and the number of clusters.

\begin{algorithm*}[htb]
  \caption{Hierarchical clustering }
\label{algo:hierarchical}
\begin{algorithmic}[1]

  \STATE Goal: given the $N$ snapshots in a $D$-dimensional space in
  the form of the matrix $X$, identify a
  clustering with $nc$ clusters, such that each snapshot is assigned to
  a cluster, by merging clusters with the highest similarity.

  \STATE Set the linkage criterion and the number of clusters, $nc$.
  
  \STATE Initialize the clusters by putting every snapshot in its own cluster.

  \STATE Compute the similarity between all pairs of clusters using
  the selected linkage criterion and distance metric.

  \STATE Merge the two clusters that are the most similar into one cluster.

  \STATE Repeat steps 4 and 5 until the desired number of $nc$ clusters
  has been obtained or the similarity between clusters no longer
  satisfies a specified level.

\end{algorithmic}
\end{algorithm*}

One of the benefits of hierarchical clustering in processing extremely
high-dimensional data is that it is not an iterative algorithm,
requiring only the pairwise distance matrix between the data points.
This allowed us to generate results using the original data set,
without the need to create an approximation using random projections.
While calculating the pairwise distance matrix is computationally
expensive, it can be computed once and then used to test the algorithm
with several different parameters.

%
\section{Results and discussion}
\label{sec:results}
%

We next present the results of clustering the snapshots for the three
variables using three methods: i) random projections followed by
k-means, ii) k-means using the weight vectors from SVD, and iii) hierarchical
clustering on the original snapshot matrix. We also discuss how we set
various parameters in the algorithms to generate a cluster assignment
for each snapshot.

All codes, unless otherwise stated, were written in C++ and Python. We
exploited parallelism where possible through the use of the sub-process
capability in Python, but did not explicitly optimize the conversion
of the HDF5 files to the snapshot matrices, which was the more time
consuming part of our solution.

\begin{figure}[bht]
\centering
\setlength\tabcolsep{1pt}
\begin{tabular}{cc}
\includegraphics[width=0.45\textwidth]{r01_i017_00000_t00_nmass.pdf} &
\includegraphics[width=0.45\textwidth]{r02_i028_00000_t00_nmass.pdf} \\
\includegraphics[width=0.45\textwidth]{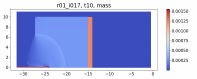} & 
\includegraphics[width=0.45\textwidth]{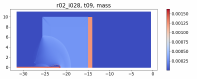} \\
\includegraphics[width=0.45\textwidth]{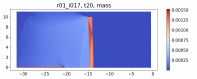} &
\includegraphics[width=0.45\textwidth]{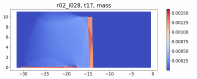} \\
\includegraphics[width=0.45\textwidth]{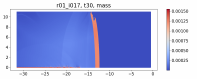} &
\includegraphics[width=0.45\textwidth]{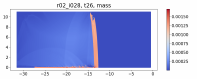} \\
\includegraphics[width=0.45\textwidth]{r01_i017_14157_t41_nmass.pdf} &
\includegraphics[width=0.45\textwidth]{r02_i028_13841_t35_nmass.pdf} \\
\end{tabular}
\caption{The variable {\it mass}, after the snapshots have been
  aligned, cropped, and remapped, at different time steps in two
  simulations showing the evolution of the data over time.  Left: key
  r01\_i017 (no break case) at time steps t0, t10, t20, t30, and t41.
  Right: key r02\_i028 (break case) at time steps t0, t09, t17, t26,
  and t35. The color bars are different between simulations and across
  time steps. }
\label{fig:timesteps_nmass}
\end{figure}

We begin by first taking a closer look at the data to be clustered,
now focusing on multiple time steps in two example simulations, key
r01\_i017 and key r02\_i028, after the original output has been
pre-processed as discussed in Section~\ref{sec:preprocessing}.
Figure~\ref{fig:timesteps_nmass} shows how the mass variable evolves
with time in these two simulations.  

We again observe that the first
and last time steps in the example simulations are very different,
suggesting there are at least two clusters in the data. In addition,
there is similarity in intermediate time steps, even though one
simulation is a break case and the other a no-break case. This
suggests that there is inherent clustering in the data, though we
expect the clusters will not be well separated as any clustering would
assign some neighboring time steps, which are very similar, to
different clusters. The corresponding images for x-momentum and
y-momentum are shown in Figures~\ref{fig:timesteps_nxmom}
and~\ref{fig:timesteps_nymom} in Appendix~\ref{sec:appendix1}
and~\ref{sec:appendix2}, respectively.  Unlike the mass variable, the
intermediate time steps are less similar between the break and
no-break case.

%
\subsection{Results with random projections and k-means clustering}
\label{sec:results_kmeans}
%

Our approach to clustering using random projections followed by
k-means clustering, introduces randomness in the projections and
in the choice of initial cluster centers, both of which can influence
the assignment of snapshots to clusters. If this influence is large,
it can result in uncertainty in cluster assignment.

%
\subsubsection{Understanding the effect of the randomness of random projections}
\label{sec:results_randproj}
%

\begin{figure}[ht]
\centering
\begin{tabular}{cc}
\includegraphics[width=0.4\textwidth]{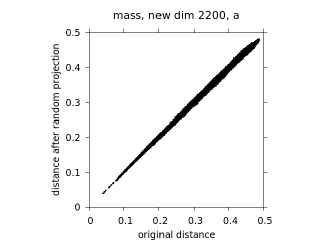} &
\includegraphics[width=0.4\textwidth]{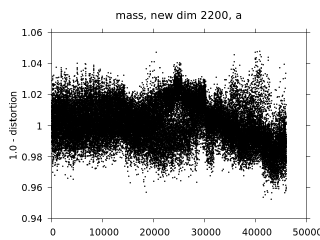} \\
\includegraphics[width=0.4\textwidth]{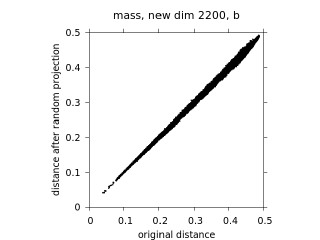} &
\includegraphics[width=0.4\textwidth]{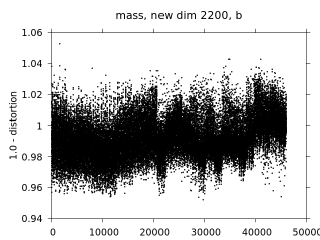} \\
\end{tabular}
\caption{For the mass variable, evaluating the choice of reduced
  dimension and randomization on the distances between the 29
  snapshots from the baseline simulation to all 1604 snapshots. Left -
  actual vs. random-projected distances for two repetitions of random
  projections indicating that the randomness of the random projections
  has little effect on the new distances. Right - the
  corresponding values of $(1.0-\epsilon)$ indicating that the
  distortion $\epsilon$ is much less than 10\%, even though our choice of
  reduced dimension of 2200 is smaller than the 6325 suggested by the
  Johnson-Lindenstrauss lemma for $\epsilon = 0.1$.  }
\label{fig:randproj_results_nmass}
\end{figure}

We first empirically evaluated the effect of the randomness in the
projection by repeating the projection of the snapshot matrix twice
for a reduced dimension $d = 2200$. This new dimension gives nearly a
1000-fold reduction from the original dimension of $D = 2,180,799$. We
calculated the original and reduced-dimensional distances between all
snapshots and the 29 snapshots of the {\it baseline} simulation with
HE length, jet tip velocity, and jet radius equal to 5.0cm,
0.950cm/$\mu$s, and 0.125cm, respectively. This simulation is at an
extreme corner of the input space. The results for the mass variable
are shown in Figure~\ref{fig:randproj_results_nmass}; the
corresponding figures for the x- and y-momentum are shown in
Figures~\ref{fig:randproj_results_nxmom}
and~\ref{fig:randproj_results_nymom} in Appendix~\ref{sec:appendix1}
and~\ref{sec:appendix2}, respectively.

These figures indicate that for our data set, the randomness of the
random projections has little effect on the distances between
snapshots, unlike the results of Fern and
Brodley~\cite{fern2003:randproj}, who found random projections to be
unstable for clustering. Though we present the results only for two
repetitions, we observed similar behavior in multiple repetitions.  In
addition, as has been observed by others~\cite{bingham2001:randproj},
we can use a smaller reduced dimension than the one suggested by the
Johnson-Lindenstrauss lemma (Equation~\ref{eqn:reddim}) for a specific
value of the distortion $\epsilon$.  Our choice of $d = 2200$, which
is much smaller than the dimension of 6325 suggested by the lemma for
$\epsilon = 0.1$, results in a distortion less than $0.05$ for most
snapshots.

%
\subsubsection{Exploiting the randomness of k-means to determine number of clusters}
\label{sec:results_selectk}
%

Next we applied the k-means algorithm to the randomly-projected snapshot
matrix. As the results depend on the random choice of the initial
centroids, we used ensemble clustering, where we repeat the algorithm
to generate results for 10 different initial cluster centroids. In our
implementation of k-means (Algorithm~\ref{algo:kmeans}), we set {\it
  niter}, the maximum number of iterations to 100 and {\it thresh},
the maximum distance moved by any centroid between iterations to be
0.0. For our data, the latter condition is satisfied before 100 iterations.

Another parameter we need to select is the number of clusters, $nc$.
A typical approach used in building spatio-temporal surrogates is to
generate results for different values of $nc$, and select the best
based on some metric. This metric can be the quality of reconstruction
of test snapshots~\cite{amsallem2012:nonlocal}, which is
computationally expensive for a large data set like ours, or silhouette
analysis~\cite{daniel2020:romnet}, which is useful when the clusters
are compact and clearly separated~\cite{rousseeuw1987:silhouette},
which is not the case for our data.  In practice, the number of
clusters is a trade-off between larger values that give a better
approximation of the non-linear manifold, and smaller values that make
the class assignment of the snapshots more stable.

To determine the number of clusters, we first tried the iterative
consensus clustering (ICC) method~\cite{meyer2013:icc}, which had its
own parameters that were difficult to set. It required generating
results with a range of values for $nc$ and combining them into a
consensus matrix, which identifies the number of times two snapshots
are clustered together.  We found the results to be sensitive to the
range of values of $nc$; they also did not give a clear indication of
the number of clusters inherent in the data. In addition, once the
number of clusters had been identified, the method did not provide a
way to obtain a cluster assignment for the snapshots.

However, we realized that a consensus matrix could be exploited to
understand the sensitivity of the clustering results to both the
number of clusters and the randomness of the random projections, and,
in addition, to identify the number of clusters. For our data set with
1604 snapshots, we were interested in moderate- to large-sized
clusters where possible.  We ran the k-means ensemble clustering for
small values of $nc = 3, 4, 5, \text{ and } 6$, with 10 repetitions
each, changing the initial centroids each time.  Unlike the ICC method
that generated a single consensus matrix combining all these results,
we generated a separate consensus matrix for each $nc$ and for each of
the two repetitions of the random projections (referred to as a and
b).  Our consensus matrices represented the fraction of times two
snapshots were in the same cluster; with matrix entries between 0.0
and 1.0, it became easier to set parameters for subsequent processing.

Using the mass variable as an example, we next describe how we analyze
the consensus matrices to identify the number of clusters and the
cluster assignment for the snapshots.
Table~\ref{tab:cmatrix_nmass_stats} shows the statistics on the eight
consensus matrices.  For $nc = 3 \text{ and } 4$, most of the values
are either 0.0 (snapshots never in same cluster) or 1.0 (snapshots
always in same cluster). However, as the number of clusters increases
to $nc = 5 \text{ and } 6$, there is a range of values, indicating
that some snapshots are clustered together only occasionally. This
suggests that for larger $nc$, the results are sensitive to the
initial cluster centroids. Further, for a given value of $nc$, there
is less difference between the results of the two random projections
when $nc = 3 \text{ and } 4$, than when $nc = 5 \text{ and } 6$, that
is, the clusters are more stable for lower $nc$. This last observation
does not necessarily hold for the x- and y-momentum variables as shown
in Tables~\ref{tab:cmatrix_nxmom_stats}
and~\ref{tab:cmatrix_nymom_stats} in Appendix~\ref{sec:appendix1}
and~\ref{sec:appendix2}, respectively.

\begin{table}[htb]
  \begin{center}
    \begin{tabular}{|l ||l|l|| l|l ||l|l ||l|l|}
      \hline
      Values & nc3, a & nc3, b & nc4, a & nc4, b & nc5, a & nc5, b & nc6, a & nc6, b \\ 
      \hline
      0.0 &  59.18 &  59.34  &  63.72  &  63.75  &  64.55  &  63.43  &  65.01  &  65.73 \\
      0.1 &  0.0   &  0.0    &  0.91   &  0.01   &  0.01   &  0.0    &  12.86  &  1.15  \\
      0.2 &  0.0   &  0.0    &  0.65   &  0.89   &  0.0    &  0.91   &  1.67   &  1.65  \\
      0.3 &  0.0   &  0.0    &  0.0    &  0.63   &  13.81  &  0.59   &  1.19   &  0.40  \\
      0.4 &  0.0   &  0.0    &  0.0    &  0.0    &  0.09   &  0.003  &  0.52   &  11.62 \\
      0.5 &  0.41  &  0.0    &  0.0    &  0.0    &  0.0    &  16.95  &  0.04   &  1.50  \\
      0.6 &  0.0   &  0.0    &  0.0    &  0.0    &  0.01   &  0.03   &  0.63   &  1.49  \\
      0.7 &  0.0   &  0.0    &  0.0    &  0.39   &  5.27   &  1.31   &  0.80   &  0.41  \\
      0.8 &  0.0   &  0.0    &  0.51   &  0.63   &  0.0    &  1.01   &  1.14   &  2.12  \\
      0.9 &  0.0   &  0.0    &  0.57   &  0.02   &  0.09   &  0.03   &  2.57   &  0.69  \\
      1.0 &  40.41 &  40.66  &  33.65  &  33.68  &  16.17  &  15.74  &  13.58  &  13.24 \\
      \hline
    \end{tabular}
  \end{center}
  \vspace{0.2cm}
  \captionof{table}{Distribution of values in the consensus matrices
    for the mass variable, with 10 repetitions of the clustering. We
    used four values for $nc$, the number of clusters, and two
    repetitions of the random projections, indicated by a and b. The
    values in the table indicate how often (in percentage) the value
    in the first column occurs in the matrix. Thus, for ($nc=3$, a) in
    the second column, nearly 60\% of the matrix entries are 0.0,
    nearly 40\% are 1.0, with a small percentage of entries at 0.5. In
    contrast, for $nc=6$, both a and b, all values between 0.0 and 1.0
    appear in the consensus matrix, and only 79\% of the entries are
    either 0 or 1, which indicates how often snapshots are never or always
    in the same cluster, respectively.}
  \label{tab:cmatrix_nmass_stats}
\end{table}
\begin{figure}[!htb]
\centering
\begin{tabular}{cc}
\includegraphics[width=0.45\textwidth]{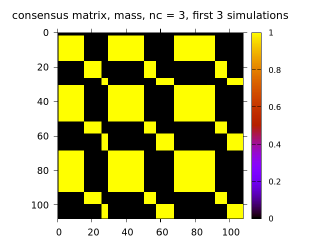} &
\includegraphics[width=0.45\textwidth]{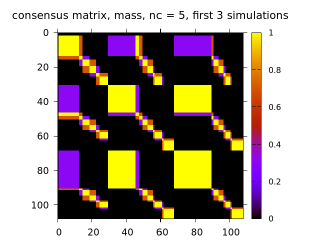} \\
\end{tabular}
\captionof{figure}{A $108 \times 108$ subset of the consensus matrix
  for mass for ($nc=3$, a) (left) and ($nc = 5$, a) (right). The
  results are for the first three simulations with 29, 38, and 41
  snapshots, respectively, shown in order, with the first snapshot of
  the first simulation at top left corner, and the 41-st snapshot of
  the third simulation at the bottom right corner.  The left figure
  indicates that the early time steps of the three simulations are in
  the same cluster, and that a small number of snapshots at late time
  in the first simulation are in the same cluster as the late time
  steps in the other two simulations. The plot on the right has a
  larger range of values, showing that the early time steps of the
  first simulation are less often in the same cluster as the early
  time steps of the second and third simulations, while the latter two
  are usually in the same cluster.  Also, the cluster assignment of
  later time steps in a simulation is less clear with smaller
  clusters.}
\label{fig:cmatrix_nmass1}
\end{figure}

Figure~\ref{fig:cmatrix_nmass1} shows a small $108 \times 108$ subset
of the consensus matrices for ($nc=3$, a) and ($nc=5$, a) that further
highlight the differences between the small and large values of $nc$.
The subsets show the first three simulations with 29, 38, and 41
snapshots, respectively, in order of the time steps, starting with the
first time step of the first simulation, and ending at the 41-st time
step of the third simulation. As described in the caption, we can
visually identify which snapshots in which simulations are in the same
cluster.

This observation gives us a simple approach to cluster assignment for
the snapshots. We start with the first row of a consensus matrix, and
put all snapshots with a value 1.0 in this row into one cluster.
We repeat with the rows corresponding to each of these snapshots, and
so on, until we have identified all snapshots in the first cluster.
Then, we identify a snapshot not yet assigned to a cluster, and repeat
to create the second cluster. And so on. This approach gives us the
class assignment for the snapshots.  However, we found that if we only
combine snapshots that are always grouped together (that is, matrix
entries with value 1.0), we will identify more clusters than $nc$
because a snapshot that occurs together with another only nine times
out of ten (with a value of 0.9 in the matrix) would be in a different
cluster.

\begin{figure}[!htb]
\centering
\begin{tabular}{cc}
\includegraphics[width=0.45\textwidth]{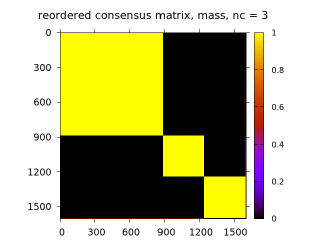} &
\includegraphics[width=0.45\textwidth]{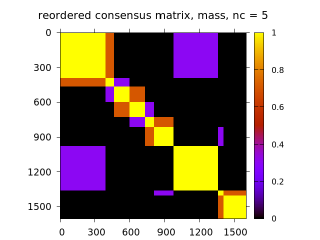} \\
\end{tabular}
\captionof{figure}{The reordered consensus matrix for mass showing all
  snapshots reordered by class. Left: matrix for (nc3, a); right: 
  matrix for (nc5, a).}
\label{fig:cmatrix_nmass2}
\end{figure}

We observed this behavior for both ($nc=3$, a) and ($nc=5$, a) as
shown in Figure~\ref{fig:cmatrix_nmass2} using the full consensus
matrix with the snapshots reordered by the cluster number. For
($nc=3$, a), we have five clusters, the three clearly seen in the
figure and two small ones with 4 and 2 snapshots that appear at the
end. In contrast, ($nc=5$, a) has nine clusters instead of 5, as
indicated by the diagonal blocks with values equal to 1.0.  If we
identified the clusters using matrix entries greater than a threshold,
say 0.7, the results were sensitive to the threshold.  We also had to
address the issue of entries with values equal to 0.5, which occurs
when neighboring time steps are at a cluster boundary and assigned to
one cluster or the other half the time.

\begin{figure}[!htb]
\centering
\begin{tabular}{cc}
\includegraphics[width=0.45\textwidth]{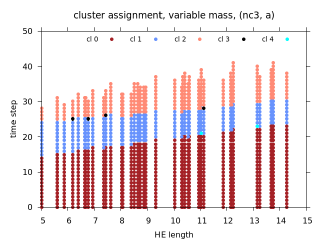} &
\includegraphics[width=0.45\textwidth]{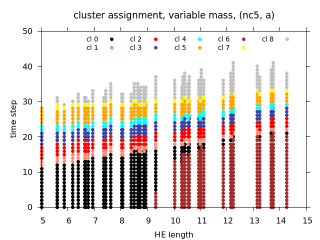} \\
\end{tabular}
\captionof{figure}{The cluster assignment for mass for the time steps
  in the simulations plot along with the HE length. The points are
  colored by their class. Left: cluster assignment for ($nc = 3$, a)
  (left) and ($nc = 5$, a) (right). In the left plot, note the two
  very small clusters, one in black with 4 snapshots and the other in
  cyan with 2 snapshots, indicating entries in the consensus matrix
  with a value of 0.5. On the right, the multiple small clusters are
  the result of a larger $nc = 5$ and the relatively large number of
  matrix entries at 0.7 and 0.3 that form their own clusters. Note
  that the simulation at the smallest HE length, which is the baseline
  simulation, does not have any time steps in cluster 8, shown in
  grey, which is of greatest interest in our problem as it has the
  late time data for the rest of the simulations.  }
\label{fig:cmatrix_nmass3}
\end{figure}

To understand how the time steps for the 45 simulations are
distributed across these 5 and 9 clusters,
Figure~\ref{fig:cmatrix_nmass3} shows the cluster assignment for each
snapshot as a function of HE length.  For ($nc=3$, a), there are two
small clusters, with 4 and 2 snapshots, that are at the time step
boundary between two clusters and could be assigned to either. For
($nc=5$, a), there are several small clusters that are connected to
other clusters either weakly or strongly, with matrix values 0.3 and
0.7, respectively, based on Table~\ref{tab:cmatrix_nmass_stats}. An
option in this specific case is to use the reordered consensus matrix
to merge a small cluster with another if they are strongly connected.
However, when the consensus matrix entries take values closer to 0.5,
it becomes less clear which clusters should be merged and why. In
addition, when a small group of consecutive (in time) snapshots in a
simulation spans 4 or 5 clusters, as shown in
Figure~\ref{fig:cmatrix_nmass3}, for ($nc=5$, a) at large values of HE
length, it is not clear whether these snapshots should remain with
their original large cluster, or be merged into the cluster of
neighboring (in time) snapshots.  We suspect that this poor cluster
assignment is the result of either too large a value of $nc$ and/or
the clustering of snapshots being more sensitive to the initial choice
of centroids for larger $nc$, resulting in values in the consensus
matrix other than 0.0, 0.5, or 1.0.

\begin{figure}[!htb]
\centering
\begin{tabular}{cc}
\multirow{3}*{\includegraphics[width=0.45\textwidth]{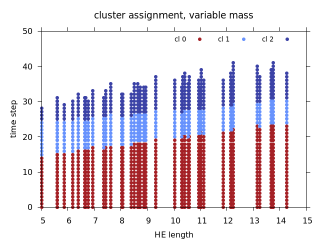}} &
 \includegraphics[width=0.45\textwidth]{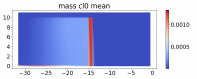} \\
& \includegraphics[width=0.45\textwidth]{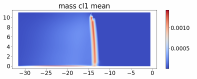} \\
& \includegraphics[width=0.45\textwidth]{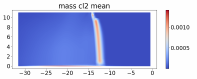} \\ 
\end{tabular}
\captionof{figure}{Left: the cluster assignment, corresponding to ($nc
  = 3$, b), selected for the mass variable shown for the times steps
  in the simulations plot along with the HE length. The cluster sizes
  for clusters 0, 1, and 2, are 889, 351, and 364, respectively.
  Right: the mean snapshot for the three clusters for the mass
  variable, from top to bottom, cluster 0, 1, and 2, that show the
  movement of the plate over time.}
\label{fig:cass_nmass1}
\centering
\begin{tabular}{cc}
\includegraphics[width=0.45\textwidth]{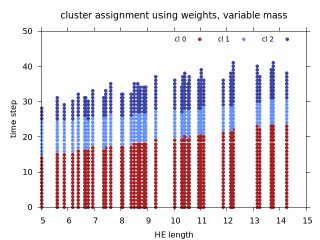} &
\includegraphics[width=0.45\textwidth]{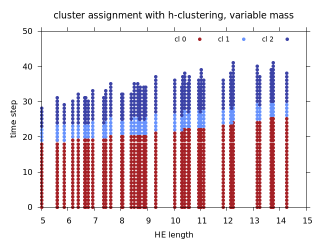} \\
\end{tabular}
\captionof{figure}{Cluster assignment for the mass variable shown for
  the times steps in the simulations plot along with the HE length.
  Left: clustering SVD weights, with $nc = 3$; the cluster sizes for
  clusters 0, 1, and 2, are 890, 352, and 362, respectively.  Right:
  using hierarchical clustering with the Ward linkage criterion, with
  $nc = 3$; the cluster sizes for clusters 0, 1, and 2, are 995, 199,
  and 410, respectively. }
\label{fig:cass_nmass2}
\end{figure}

This analysis suggests that we should select the number of clusters
that gives more repeatable results in the ensemble as it reflects a
more stable clustering. Using this criterion, we chose the cluster
assignment identified by ($nc = 3$, b) for the mass variable as it has
only 0.0 and 1.0 in the consensus matrix; the result is shown in
Figure~\ref{fig:cass_nmass1}.  The results for the x- and y-momentum
variables, using random projections and k-means clustering, are shown
in Appendix~\ref{sec:appendix1} and~\ref{sec:appendix2}, respectively,
and include the distribution of values in the consensus matrix
(Tables~\ref{tab:cmatrix_nxmom_stats}
and~\ref{tab:cmatrix_nymom_stats}), along with the cluster assignment
(Figures~\ref{fig:cass_nxmom1} and~\ref{fig:cass_nymom1}). These
results indicate 4 clusters for the x- and y-momentum variables. We
discuss these results further in Section~\ref{sec:discussion}.

%
\subsection{Results with k-means clustering using weights from SVD}
\label{sec:results_kmeans_clwt}
%

Next, we applied k-means to the weights that formed the reduced
representation of each snapshot obtained using the SVD of the full
snapshot matrix as described in Section~\ref{sec:wtclustering}. We
used the number of clusters that were identified using random
projections with k-means (Section~\ref{sec:results_selectk}) for the
three variables. As a sanity check, we repeated the clustering ten
times for each variable and confirmed that the clustering results
presented in Figures~\ref{fig:cass_nmass2},~\ref{fig:cass_nxmom2},
and~\ref{fig:cass_nymom2} for the mass, x- and y- momentum,
respectively, were stable. These results are discussed further in
Section~\ref{sec:discussion}.

%
\subsection{Results with hierachical clustering}
\label{sec:results_hierarchical}
%

We used the implementation of hierarchical clustering available in the
SciPy Python library~\cite{scipy2020:article}, version 1.10.1.  This
method requires three parameters: the number of clusters, a linkage
criterion, and a distance metric between two snapshots.  We considered
both three and four clusters for each of the three variables as these
were the most likely values for $nc$ identified using the consensus
matrix analysis (Section~\ref{sec:results_selectk}). We wanted to
evaluate whether hierarchical clustering on the original matrix $\bf
X$ would give different results from the k-means method.  To allow
experimentation with different parameters, we precomputed the pairwise
distance matrix using Euclidean distances. We used a serial
implementation for the distance calculation, though the block form of
the snapshot matrix (Equation~\ref{eqn:blockx}) could be exploited for
parallel computation. As there is no randomness in hierarchical
clustering, we ran the method once.

SciPy provides several linkage criteria that define the similarity (or
dissimilarity) between clusters; at each step, the two clusters that
are the most similar, or least dissimilar, are merged.  For the
single, complete, and average linkage criteria, the dissimilarity
between two clusters is defined as the minimum, maximum, and average
Euclidean distance, respectively, between two snapshots, one in each
of the two clusters.  The Ward linkage defines the dissimilarity
between two clusters as the increase in the error sum of squares
($ESS$) when the two clusters are merged. If ${\bf x}_i$ is a snapshot
in a cluster with $n$ snapshots, whose mean is $\bar{\bf x}$, then
$ESS$ of a cluster is defined as:
\begin{equation}
\label{eqn:ESS}
  ESS = \Sigma_{i=1}^{n} ({\bf x}_i - \bar{\bf x})^2 .
\end{equation}
Note the similarity between the definition of Ward linkage and the
k-means algorithm, where each snapshot is assigned to the closest
centroid.

We evaluated the different linkage criteria using two quality metrics.
First, we wanted the clusters to be of moderate size, especially the
cluster at late time that was of interest in our problem. Second, as
neighboring time steps in a simulation are similar, we wanted all the
snapshots in each simulation that are assigned to a cluster to be at
contiguous time steps. We observed the following:

\begin{itemize}

\item As neighboring time steps in a simulation are similar, we
  expected the single linkage criterion would place
  all or most snapshots within a single cluster. We found this to be
  true for all three variables.

\item The complete linkage criterion on the mass variable created a
  small cluster of late time snapshots, but only for the simulations
  with moderate to large HE length, excluding similar snapshots at
  late time steps for small HE length. For the x-momentum variable,
  complete linkage split some clusters across non-contiguous time
  steps in a simulation, while results for the y-momentum variable
  were acceptable.

\item Average linkage for the mass variable gave results similar to
  those with complete linkage, but performed poorly for the other two
  variables.  For the x-momentum variable, with $nc = 3$, it created a
  single cluster composed of the late time snapshots in two simulations at
  small HE length; for $nc = 4$, this cluster was split into
  two, one for each simulation.  For the y-momentum variable, with $nc
  = 3$, some simulations had all snapshots assigned to one cluster,
  while for $nc = 4$, the early and late-time snapshots were merged into a
  single cluster, while the mid-time snapshots were in different
  clusters.

\item Ward linkage generated the most meaningful cluster assignments.
  These are shown in
  Figures~\ref{fig:cass_nmass2},~\ref{fig:cass_nxmom2},
  and~\ref{fig:cass_nymom2} for the mass, x- and y- momentum
  variables, respectively, and discussed further in
  Section~\ref{sec:discussion}.

\end{itemize}

%
\subsection{Discussion}
\label{sec:discussion}
%

We next discuss the cluster assignment results from the three
clustering methods shown in Figures~\ref{fig:cass_nmass1}
and~\ref{fig:cass_nmass2} for the mass variable,
Figures~\ref{fig:cass_nxmom1} and~\ref{fig:cass_nxmom2} in
Appendix~\ref{sec:appendix1} for the x-momentum variable, and
Figures~\ref{fig:cass_nymom1} and~\ref{fig:cass_nymom2} in
Appendix~\ref{sec:appendix2} for the y-momentum variable. These
figures also include the mean snapshots for each cluster generated
using the cluster assignment from the k-means algorithm with random
projections.  These results, and our experiences, indicate the
following:

\begin{itemize}

\item We found that the theory behind random projections and
  the Johnson-Lindenstrauss lemma works in practice. As observed by
  others~\cite{bingham2001:randproj}, it is possible to obtain low
  distortion in the projected data even though the reduced
  dimension is smaller than the value recommended for a specific
  distortion. For our data set, we obtained a thousand-fold reduction in
  dimension with $\approx$5\% distortion of the data.

\item We can optimize the calculation of the randomly-projected data
  by using a very sparse random projection
  matrix~\cite{li2006:randproj} and reordering the computation such
  that the projected matrix is generated incrementally by reading in
  the data one row at a time and generating the random matrix on the
  fly, a column at a time. This initial step required for the k-means
  method takes roughly the same time as the calculation of the
  distance matrix for the hierarchical clustering, making the two
  methods competitive.

\item The results obtained by clustering using k-means and the weights
  from the SVD of the snapshot matrix are close to those obtained
  using random projections and k-means. This indicates that the
  approximation resulting from random projections has little effect on
  the clustering results.

\item Selecting a value for $nc$, the number of clusters, and
  generating a class assignment for the snapshots, remains
  challenging. For k-means, we used the randomness in choice of
  initial centroids and random projections, as well as the structure
  in the consensus matrix, to identify a stable clustering. For our
  data, we found smaller values of $nc$ gave more consistent results,
  but the similarity between snapshots at consecutive time steps
  resulted in uncertainty in cluster assignments for some snapshots.
  We could exploit domain information to assign such snapshots in very
  small clusters to other larger clusters, but this may not be an
  option if there are too many small clusters and it is not clear
  which clusters should be merged.  For hierarchical clustering, an
  analysis of the dendrograms did not shed any light on the
  appropriate number of clusters; we therefore used the number of
  clusters obtained from random projections and k-means clustering.

\item Hierarchical clustering is sensitive to the choice of the
  linkage criterion, which is expected. As has long been
  observed~\cite{jain88:clusbook}, the Ward linkage produces the best
  results, generating a meaningful cluster assignment for the time
  steps in a simulation. This might be expected, given the similarity
  between the Ward linkage and the k-means algorithm.

\item The plots of the cluster assignment for the snapshots and the
  HE length helped to evaluate whether the results are meaningful. As
  expected, each cluster captures the behavior over contiguous (in
  time) snapshots, with the early time snapshots in one cluster, the
  late time in another, and the mid-time snapshots in one or more
  clusters. This structure in the plots, where each cluster, assigned
  a different color, appears as a band, is clearer for the mass and
  y-momentum variables, but less so for the x-momentum, especially
  with hierarchical clustering, where the banded structure is less
  well defined (Figure~\ref{fig:cass_nxmom2}).

\item Comparing the clustering results for the three variables, we see
  that the assignment of snapshots to clusters, as well as the sizes
  of the clusters, are very different. 

  For the mass variable, the early time cluster is the largest. For
  the two methods based on k-means, the mid- and late-time clusters
  are of the same size, while hierarchical clustering gives a smaller
  mid-time cluster and larger early- and late-time clusters. Each
  simulation has snapshots in all three clusters.

  This is not the case for the x-momentum variable, where the
  simulations with small HE length have time steps assigned to just
  three of the four clusters.  This is because the cluster to which
  the very early time steps are assigned depends on the value of HE
  length. At early time, the x-momentum variable is dominated by the
  jet that enters from the left of the domain, moving to the right.
  Since we cropped the simulations with larger HE length on the left
  during pre-processing, the jet is not seen in these simulations in
  the very early time steps, which are therefore assigned to a
  different cluster.

  For the y-momentum variable, we observe two unusual aspects of the
  cluster assignment.  First,  all clustering methods create
  a relatively small cluster just after early time; this cluster is
  wedge shaped, thin for small HE length and growing wider as the HE
  length increases (Figures~\ref{fig:cass_nymom1}
  and~\ref{fig:cass_nymom2}). This cluster captures what happens as
  the wave that emanates from the tip of the jet
  (Figure~\ref{fig:timesteps_nymom}) reaches the top of the domain.
  Second, when we use the k-means method, a small number of snapshots
  at low values of HE length, which appear to be part of the
  wedge-shaped cluster, are in fact assigned to the late time cluster,
  which is now split across time steps. This result was consistent
  across multiple runs of the k-means ensemble, for clustering using
  both the weights from the SVD and the randomly-projected snapshot
  matrix with runs a and b, though the apparently mis-assigned
  snapshots varied slightly.  We suspect that these snapshots may
  indeed be closer to the centroid of the late-time cluster, if only
  by a tiny amount.  Changing the cluster assignment of these 13
  snapshots to avoid splitting the late-time cluster across time steps
  resulted in minor changes in the mean snapshots of the corresponding
  clusters (Figure~\ref{fig:cass_nymom3}); we selected this as the
  cluster assignment for the y-momentum variable. Note that the
  hierarchical clustering does not split the late time cluster.

\item Comparing the different clustering methods, we find that
  clustering using k-means, with either the random-projected snapshots
  or the weights from SVD, works quite well. It is a simple method,
  and though it was proposed several decades ago, it is still an
  admissible method~\cite{jain2010:kmeans}. As it is iterative, issues
  such as a poor initial choice of centroids, or a snapshot assigned
  to the wrong cluster, are fixed in later iterations, unlike
  hierarchical clustering. For our data set, with the high-dimensional
  snapshot matrix in block form, k-means requires the reduction in
  dimension through random projection, while hierarchical clustering
  requires calculation of the distance matrix; both take roughly the
  same compute time. We were able to use the randomness of the initial
  centroids in k-means to select the number of clusters, $nc$, but
  there was no such option we could use for hierarchical clustering
  with Ward linkage.

\end{itemize}

%
\section{Conclusions}
\label{sec:conc}
%

In this report, we used output from simulations of a jet interacting
with high explosives to address two challenges in building
spatio-temporal surrogates for high-dimensional data. First, the data
were available on spatial domains of different sizes, at grid points
that varied in their spatial coordinates, and in a format that
distributed the output across multiple files at each time step of the
simulation. We described how we reorganized these large data sets into
a consistent format efficiently, exploiting parallelism where
possible.  Second, to improve the accuracy of the surrogates, we
wanted to cluster the data by similarity and build a separate
surrogate for each cluster. However, as the outputs are
high-dimensional, with the spatial domain represented by more than two
million grid points, traditional iterative clustering methods, such as
k-means, could not be applied directly.  We showed how we could use
random projections to make the clustering of these outputs tractable.
Our experiences indicated that the approximation introduced by the
random projections had little effect on the clustering results.
Moreover, we could use the randomness of both the random projections
and the initial choice of cluster centroids in k-means to identify the
number of clusters.  The effectiveness of our approach is discussed
further in the companion report~\cite{kamath2023:stmapps}, where we
show how clustering the data by similarity improves the accuracy of
the spatio-temporal surrogates created from a small set of
simulations.

%
\section{Acknowledgment}
\label{sec:ack}
%

We would like to thank the Defense Threat Reduction Agency (DTRA) for
funding this work. The simulations of the interaction of the jet with
high explosives were performed using the ARES code developed at
Lawrence Livermore National Laboratory.

LLNL-TR-850159 This work performed under the auspices of the U.S. Department of
Energy by Lawrence Livermore National Laboratory under Contract
DE-AC52-07NA27344. 

This document was prepared as an account of work sponsored by an
agency of the United States government. Neither the United States
government nor Lawrence Livermore National Security, LLC, nor any of
their employees makes any warranty, expressed or implied, or assumes
any legal liability or responsibility for the accuracy, completeness,
or usefulness of any information, apparatus, product, or process
disclosed, or represents that its use would not infringe privately
owned rights. Reference herein to any specific commercial product,
process, or service by trade name, trademark, manufacturer, or
otherwise does not necessarily constitute or imply its endorsement,
recommendation, or favoring by the United States government or
Lawrence Livermore National Security, LLC. The views and opinions of
authors expressed herein do not necessarily state or reflect those of
the United States government or Lawrence Livermore National Security,
LLC, and shall not be used for advertising or product endorsement
purposes.

\bibliographystyle{plainnat}
\bibliography{ms_arxiv}

\clearpage

\appendix

%
\section{Appendix: Data and results for x-momentum}
\label{sec:appendix1}
%

\vspace{2cm}

\begin{figure}[htb]
\centering
\setlength\tabcolsep{1pt}
\begin{tabular}{cc}
\includegraphics[width=0.45\textwidth]{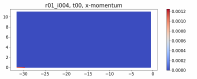} &
\includegraphics[width=0.45\textwidth]{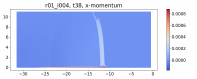} \\
\includegraphics[width=0.45\textwidth]{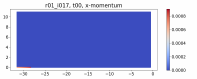} &
\includegraphics[width=0.45\textwidth]{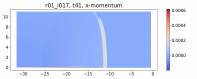} \\
\includegraphics[width=0.45\textwidth]{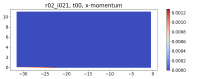} &
\includegraphics[width=0.45\textwidth]{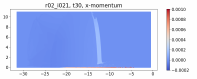} \\
\includegraphics[width=0.45\textwidth]{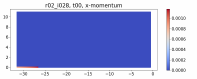} &
\includegraphics[width=0.45\textwidth]{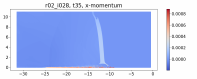} \\
\end{tabular}
\caption{The variable {\it x-momentum}, after the snapshots have been
  aligned, cropped, and remapped, at the first time step (left column)
  and last time step (right column) for the four example simulations
  in Table~\ref{tab:sample_params}. From top to bottom, simulations
  with keys r01\_i004, r01\_i017, r02\_i021, r02\_i028, respectively.
  The first time step has value zero for most of the domain, except
  the jet along the bottom on the left side of the region. The color
  bars are different between simulations and across time steps. }
\label{fig:sample_aligned_nxmom}
\end{figure}
 

\begin{figure}[htb]
\centering
\setlength\tabcolsep{1pt}
\begin{tabular}{cc}
\includegraphics[width=0.45\textwidth]{r01_i017_00000_t00_nxmom.pdf} &
\includegraphics[width=0.45\textwidth]{r02_i028_00000_t00_nxmom.pdf} \\
\includegraphics[width=0.45\textwidth]{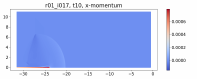} & 
\includegraphics[width=0.45\textwidth]{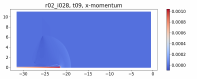} \\
\includegraphics[width=0.45\textwidth]{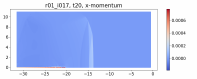} &
\includegraphics[width=0.45\textwidth]{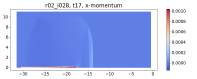} \\
\includegraphics[width=0.45\textwidth]{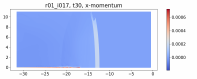} &
\includegraphics[width=0.45\textwidth]{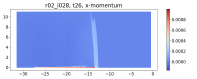} \\
\includegraphics[width=0.45\textwidth]{r01_i017_14157_t41_nxmom.pdf} &
\includegraphics[width=0.45\textwidth]{r02_i028_13841_t35_nxmom.pdf} \\
\end{tabular}
\caption{The variable {\it x-momentum}, after the snapshots have been
  aligned, cropped, and remapped, at different time steps in two
  simulations showing the evolution of the data over time.  Left: key
  r01\_i017 (no break case) at time steps t0, t10, t20, t30, and t41.
  Right: key r02\_i028 (break case) at time steps t0, t09, t17, t26,
  and t35. The color bars are different between simulations and across
  time steps.}
\label{fig:timesteps_nxmom}
\end{figure}

\begin{figure}[htb]
\centering
\begin{tabular}{cc}
\includegraphics[width=0.4\textwidth]{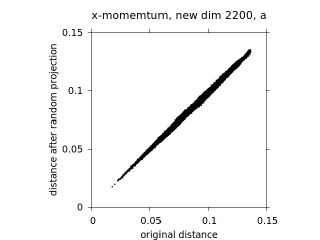} &
\includegraphics[width=0.4\textwidth]{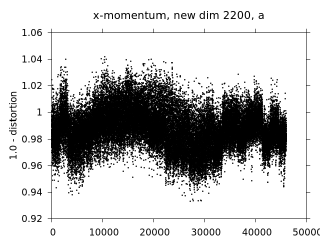} \\
\includegraphics[width=0.4\textwidth]{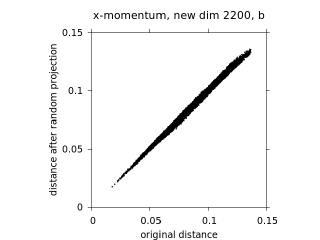} &
\includegraphics[width=0.4\textwidth]{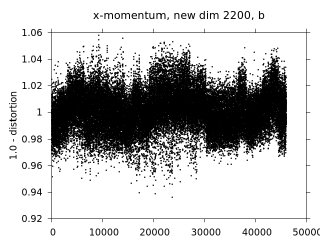} \\
\end{tabular}
\caption{For the x-momentum variable, evaluating the choice of reduced
  dimension and randomization on the distances between the 29
  snapshots from the baseline simulation to all 1604 snapshots. Left -
  actual vs. random-projected distances for two repetitions of random
  projections indicating that the randomness of the random projections
  has little effect on the new distances. Right - the corresponding
  values of $(1.0-\epsilon)$ indicating that the distortion $\epsilon$
  is less than 10\%, even though our choice of reduced dimension of
  2200 is smaller than the 6325 suggested by the Johnson-Lindenstrauss
  lemma for $\epsilon = 0.1$.  }
\label{fig:randproj_results_nxmom}
%
  \begin{center}
    \begin{tabular}{|l ||l|l|| l|l ||l|l ||l|l|}
      \hline
      Values & nc3, a & nc3, b & nc4, a & nc4, b & nc5, a & nc5, b & nc6, a & nc6, b \\ 
      \hline
      0.0 &   58.39 &  63.12 &    71.34 &    71.52 &   70.62 &   71.45 &   70.43 &   68.87 \\
      0.1 &   0.12  &  0.008 &    0.003 &    0.0   &   0.08  &   0.34  &   3.00  &   3.54  \\
      0.2 &   0.07  &  0.0   &    2.79  &    0.0   &   1.39  &   0.34  &   2.27  &   3.61  \\
      0.3 &   2.37  &  0.075 &    0.12  &    0.0   &   2.85  &   2.21  &   2.78  &   2.87  \\
      0.4 &   3.25  &  2.76  &    0.19  &    0.0   &   0.0   &   1.32  &   2.78  &   2.66  \\
      0.5 &   4.46  &  0.47  &    0.0   &    6.27  &   8.76  &   7.98  &   2.64  &   2.42  \\
      0.6 &   3.13  &  3.10  &    0.20  &    0.0   &   0.006 &   1.31  &   2.64  &   2.14  \\
      0.7 &   2.56  &  0.089 &    0.12  &    0.0   &   2.81  &   2.13  &   2.71  &   3.12  \\
      0.8 &   0.16  &  0.0005&    2.52  &    0.0   &   1.56  &   0.51  &   2.63  &   3.37  \\
      0.9 &   0.23  &  0.024 &    0.008 &    0.0   &   0.16  &   0.33  &   2.52  &   3.36  \\
      1.0 &   25.26 &  30.35 &    22.71 &    22.22 &   11.76 &   12.07 &   5.61  &   4.04  \\
      \hline
    \end{tabular}
  \end{center}
  \vspace{-0.2cm}
  \captionof{table}{Distribution of values in the consensus matrix
    generated for the x-momentum variable, with 10 repetitions of the
    clustering. We used four values for $nc$, the number of clusters,
    along with two repetitions of the random projections, indicated by
    a and b. The values in the table indicate how often the value in
    the first column occurs in the matrix. Thus, for ($nc=4$, b) in
    the fifth column, nearly 72\% of the matrix entries are 0.0,
    nearly 22\% are 1.0, with a small percentage of entries at 0.5.  }
  \label{tab:cmatrix_nxmom_stats}
\end{figure}

\begin{figure}[htb]
\centering
\begin{tabular}{cc}
\multirow{3}*{\includegraphics[width=0.45\textwidth]{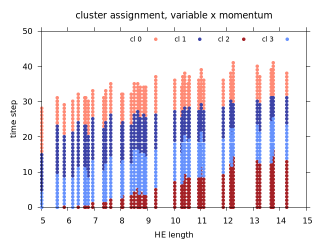}} &
 \includegraphics[width=0.45\textwidth]{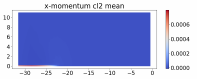} \\
& \includegraphics[width=0.45\textwidth]{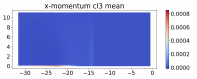} \\
& \includegraphics[width=0.45\textwidth]{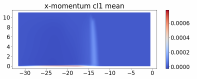} \\ 
& \includegraphics[width=0.45\textwidth]{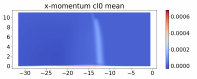} \\ 
\end{tabular}
\captionof{figure}{Left: the cluster assignment, corresponding to
  ($nc=4$, a), selected for the x-momentum variable shown for the
  times steps in the simulations plot along with the HE length.  The
  cluster sizes for clusters 0, 1, 2, and 3, are 397, 454, 305, and
  448, respectively. Right: the mean snapshot for the four clusters
  for the x-momentum variable, from top to bottom, clusters 2, 3, 1,
  and 0, showing the evolution with time. Note that the clusters form
  four bands that go across the HE length. }
\label{fig:cass_nxmom1}
\centering
\begin{tabular}{cc}
\includegraphics[width=0.45\textwidth]{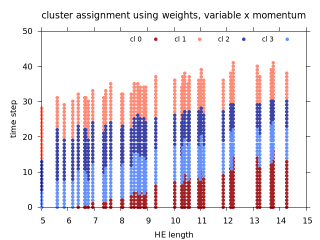} &
\includegraphics[width=0.45\textwidth]{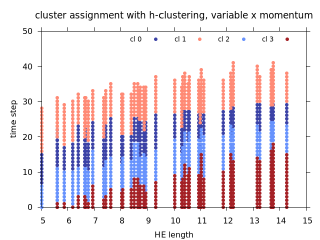} \\
\end{tabular}
\captionof{figure}{Cluster assignment for the x-momentum variable
  shown for the times steps in the simulations plot along with the HE
  length. Left: clustering SVD weights, with $nc = 4$; the cluster sizes
  for clusters 0, 1, 2, and 3, are 298, 441, 453, and 412,
  respectively.  Right: using hierarchical clustering with the Ward
  linkage criterion, with nc = 4; the cluster sizes for clusters 0, 1,
  2, and 3 are 342, 475, 404, and 383, respectively. The cluster
  number associated with a cluster of a specific color is different in
  the two plots. Note that the bands formed by the four clusters are
  less clear for hierarchical clustering.}
\label{fig:cass_nxmom2}
\end{figure}

\clearpage

%
\section{Appendix:  Data and results for y-momentum}
\label{sec:appendix2}
%

\vspace{2cm}

\begin{figure}[htb]
\centering
\setlength\tabcolsep{1pt}
\begin{tabular}{cc}
\includegraphics[width=0.45\textwidth]{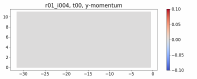} &
\includegraphics[width=0.45\textwidth]{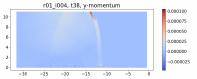} \\
\includegraphics[width=0.45\textwidth]{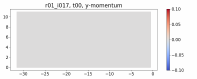} &
\includegraphics[width=0.45\textwidth]{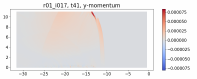} \\
\includegraphics[width=0.45\textwidth]{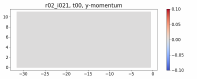} &
\includegraphics[width=0.45\textwidth]{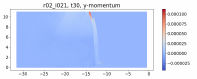} \\
\includegraphics[width=0.45\textwidth]{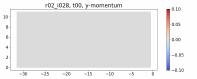} &
\includegraphics[width=0.45\textwidth]{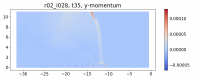} \\
\end{tabular}
\caption{The variable {\it y-momentum}, after the snapshots have been
  aligned, cropped, and remapped, at the first (left column) and last
  (right column) time steps for the four example simulations in
  Table~\ref{tab:sample_params}. From top to bottom, simulations with
  keys r01\_i004, r01\_i017, r02\_i021, r02\_i028, respectively. The
  first time step has value zero for most of the domain. Unlike the
  x-momentum in Figure~\ref{fig:sample_aligned_nxmom}, there is no
  y-momentum for the jet at initial time. The color bars are different
  between simulations and across time steps.}
\label{fig:sample_aligned_nymom}
\end{figure}

\begin{figure}[htb]
\centering
\setlength\tabcolsep{1pt}
\begin{tabular}{cc}
\includegraphics[width=0.45\textwidth]{r01_i017_00000_t00_nymom.pdf} &
\includegraphics[width=0.45\textwidth]{r02_i028_00000_t00_nymom.pdf} \\
\includegraphics[width=0.45\textwidth]{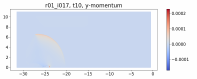} & 
\includegraphics[width=0.45\textwidth]{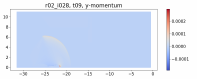} \\
\includegraphics[width=0.45\textwidth]{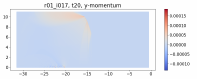} &
\includegraphics[width=0.45\textwidth]{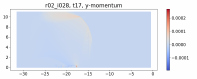} \\
\includegraphics[width=0.45\textwidth]{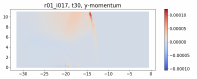} &
\includegraphics[width=0.45\textwidth]{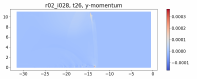} \\
\includegraphics[width=0.45\textwidth]{r01_i017_14157_t41_nymom.pdf} &
\includegraphics[width=0.45\textwidth]{r02_i028_13841_t35_nymom.pdf} \\
\end{tabular}
\caption{The variable {\it y-momentum}, after the snapshots have been
  aligned, cropped, and remapped, at different time steps in two
  simulations showing the evolution of the data over time.  Left: key
  r01\_i017 (no break case) at time steps t0, t10, t20, t30, and t41.
  Right: key r02\_i028 (break case) at time steps t0, t09, t17, t26,
  and t35. The color bars are different between simulations and across
  time steps.}
\label{fig:timesteps_nymom}
\end{figure}

\begin{figure}[htb]
\centering
\begin{tabular}{cc}

\includegraphics[width=0.4\textwidth]{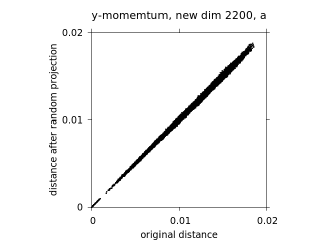} &
\includegraphics[width=0.4\textwidth]{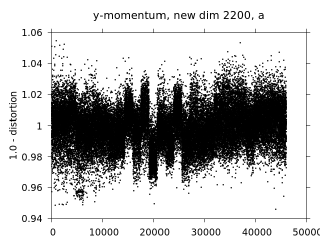} \\
\includegraphics[width=0.4\textwidth]{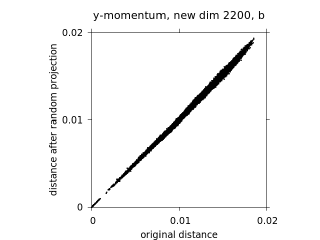} &
\includegraphics[width=0.4\textwidth]{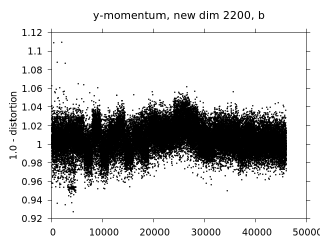} \\
\end{tabular}
\caption{For the y-momentum variable, evaluating the choice of reduced
  dimension and randomization on the distances between the 29
  snapshots from the baseline simulation to all 1604 snapshots. Left -
  actual vs. random-projected distances for two repetitions of random
  projections indicating that the randomness of the random projections
  has little effect on the new distances. Right - the corresponding
  values of $(1.0-\epsilon)$ indicating that the distortion $\epsilon$
  is less than 10\%, even though our choice of reduced dimension of
  2200 is smaller than the 6325 suggested by the Johnson-Lindenstrauss
  lemma for $\epsilon = 0.1$.  }
\label{fig:randproj_results_nymom}
  \begin{center}
    \begin{tabular}{|l ||l|l|| l|l ||l|l ||l|l|}
      \hline
      Values & nc3, a & nc3, b & nc4, a & nc4, b & nc5, a & nc5, b & nc6, a & nc6, b \\ 
      \hline
      0.0 &   50.30 &   49.53 &   69.88 &   70.65 &   66.64 &    72.12 &   71.85 &   69.97 \\
      0.1 &   0.29  &   0.62  &   0.001 &   0.11  &   1.44  &    1.32  &   0.33  &   1.27  \\
      0.2 &   0.50  &   0.008 &   0.57  &   0.0   &   1.78  &    0.99  &   1.95  &   2.19  \\
      0.3 &   0.0   &   0.10  &   0.66  &   0.0   &   1.74  &    0.82  &   1.03  &   1.03  \\
      0.4 &   14.32 &   0.92  &   0.0   &   0.0   &   2.77  &    0.009 &   0.52  &   1.54  \\
      0.5 &   1.47  &   22.31 &   0.0   &   0.0   &   0.68  &    0.06  &   4.65  &   3.62  \\
      0.6 &   7.37  &   0.75  &   0.0   &   0.0   &   1.88  &    0.03  &   2.09  &   2.18  \\
      0.7 &   0.0   &   0.16  &   0.25  &   0.0   &   3.88  &    0.68  &   1.59  &   2.12  \\
      0.8 &   0.21  &   0.06  &   0.47  &   0.0   &   2.16  &    1.38  &   1.72  &   1.31  \\
      0.9 &   0.58  &   0.30  &   0.003 &   0.18  &   1.52  &    0.94  &   0.48  &   1.82  \\
      1.0 &   24.95 &   25.23 &   28.17 &   29.06 &   15.52 &    21.65 &   13.79 &   12.93 \\
      \hline
    \end{tabular}
  \end{center}
  \vspace{-0.2cm}
  \captionof{table}{Distribution of values in the consensus matrix
    generated for the {\it y-momentum} variable, with 10 repetitions
    of the clustering. We used four values for $nc$, the number of
    clusters, along with two repetitions of the random projections,
    indicated by a and b. The values in the table indicate how often
    the value in the first column occurs in the matrix. Thus, for
    ($nc=4$, b) in the fifth column, nearly 71\% of the matrix entries
    are 0.0, nearly 29\% are 1.0, with a small percentage of entries
    at 0.1 and 0.9 }
  \label{tab:cmatrix_nymom_stats}
\end{figure}

\begin{figure}[htb]
\centering
\begin{tabular}{cc}
\multirow{3}*{\includegraphics[width=0.45\textwidth]{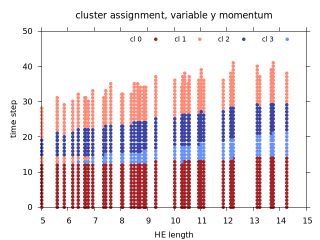}} &
 \includegraphics[width=0.45\textwidth]{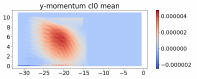} \\
& \includegraphics[width=0.45\textwidth]{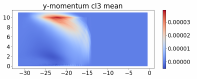} \\
& \includegraphics[width=0.45\textwidth]{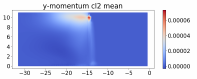} \\ 
& \includegraphics[width=0.45\textwidth]{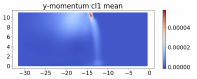} \\ 
\end{tabular}
\captionof{figure}{Left: the cluster assignment, corresponding to
  ($nc=4$, b), selected for the {\it y-momentum} variable shown for
  the time steps in the simulations plot along with the HE length.
  The cluster sizes for clusters 0, 1, 2, and 3, are 615, 460,
  371, and 158, respectively. Right: the mean snapshot for the four
  clusters for the y-momentum variable, from top to bottom, clusters
  0, 3, 2, and 1, showing the evolution with time. Note that the
  clusters form four bands across the HE length, with the much thinner
  band having some time steps at low HE length that belong to a
  different cluster; this is fixed in Figure~\ref{fig:cass_nymom3}.  }
\label{fig:cass_nymom1}
\centering
\begin{tabular}{cc}
\includegraphics[width=0.45\textwidth]{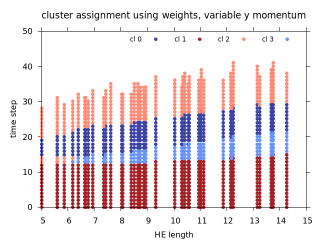} &
\includegraphics[width=0.45\textwidth]{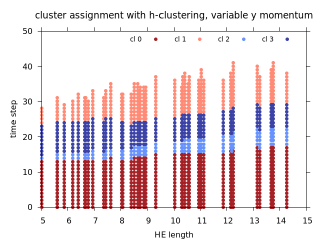} \\
\end{tabular}
\captionof{figure}{Cluster assignment for the x-momentum variable
  shown for the times steps in the simulations plot along with the HE
  length. Left: clustering SVD weights, with $nc = 4$; the cluster sizes
  for clusters 0, 1, 2, and 3, are 355, 619, 455, 175, respectively.
  Right: using hierarchical clustering with the Ward linkage
  criterion, with nc = 4; the cluster sizes for clusters 0, 1, 2, and
  3 are 697, 412, 152, and 343, respectively.  The cluster number
  associated with a cluster of a specific color is different in the
  two plots. Note that, unlike k-means, hierarchical clustering does
  not include some mid-time snapshots in the late time cluster at low
  values of HE length. }
\label{fig:cass_nymom2}
\end{figure}

\begin{figure}[htb]
\centering
\begin{tabular}{cc}
  \multirow{3}*{\includegraphics[width=0.45\textwidth]{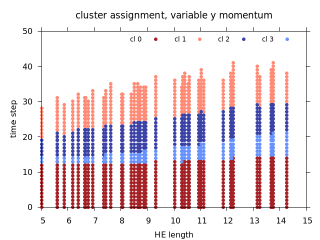}} &
 \includegraphics[width=0.45\textwidth]{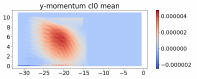} \\
& \includegraphics[width=0.45\textwidth]{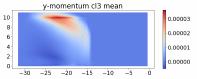} \\
& \includegraphics[width=0.45\textwidth]{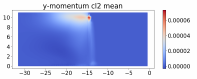} \\ 
& \includegraphics[width=0.45\textwidth]{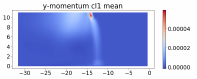} \\ 
\end{tabular}
\captionof{figure}{Left: the {\bf fixed} cluster assignment,
  corresponding to ($nc=4$, b), selected for the {\it y-momentum}
  variable shown for the time steps in the simulations plot along with
  the HE length.  The cluster sizes for clusters 0, 1, 2, and 3,
  are 615, 447, 371, and 171, respectively. 13 snapshots were moved
  from cluster 1 to cluster 3, resulting in a slight difference in
  values to the left of the plate near the top compared to the results
  in Figure~\ref{fig:cass_nymom1}. Right: the mean snapshot for the
  four clusters for the y-momentum variable, from top to bottom,
  clusters cl0, cl3, cl2, cl1, showing the evolution with time.}
\label{fig:cass_nymom3}
\end{figure}

\end{document}